\pdfoutput=1
\documentclass[11pt,letterpaper,logo,onecolumn,copyright]{template_from_google}
\usepackage[authoryear,sort&compress,round]{natbib}

\usepackage{amsmath,amsfonts,bm}

\def\eqref#1{\cref{#1}}

\def\1{\bm{1}}

\def\rvx{{\mathbf{x}}}

\DeclareMathAlphabet{\mathsfit}{\encodingdefault}{\sfdefault}{m}{sl}
\SetMathAlphabet{\mathsfit}{bold}{\encodingdefault}{\sfdefault}{bx}{n}

\usepackage{amsmath, amssymb, amsthm, mathtools}
\usepackage{bm}
\usepackage{hyperref}
\usepackage{url}
\usepackage{booktabs}
\usepackage{multirow}
\usepackage{float}
\usepackage{wrapfig}
\usepackage[capitalise]{cleveref}

\crefname{figure}{Fig.}{Figs.}
\crefname{equation}{Eq.}{Eqs.}
\crefname{table}{Tab.}{Tabs.}
\crefname{section}{Sec.}{Secs.}
\crefname{subsection}{Sec.}{Secs.}
\crefname{subsubsection}{Sec.}{Secs.}
\crefname{appendix}{Appendix}{Appendices}
\crefname{subappendix}{Appendix}{Appendices}
\crefname{subsubappendix}{Appendix}{Appendices}
\Crefname{appendix}{Appendix}{Appendices}
\Crefname{subappendix}{Appendix}{Appendices}
\Crefname{subsubappendix}{Appendix}{Appendices}

\newcommand{\bu}{\mathbf{u}}
\newcommand{\bv}{\mathbf{v}}
\newcommand{\bx}{\mathbf{x}}
\newcommand{\bff}{\mathbf{f}}
\newcommand{\br}{\mathbf{r}}
\newcommand{\bs}{\mathbf{s}}
\newcommand{\beps}{\bm{\varepsilon}}

\newcommand{\spara}[1]{\noindent\textbf{#1.}}

\newtheorem{theorem}{Theorem}
\newtheorem{lemma}[theorem]{Lemma}
\newtheorem{proposition}[theorem]{Proposition}
\newtheorem{remark}[theorem]{Remark}

\title{VA-TDM: A Unified View for Improving Few-Step Diffusion Models by Velocity Adversarial Distillation}

\makeatletter
\def\@author{{\Authfont Yihong Luo\textsuperscript{1}, Tianyu Pang, and Tianyang Hu\textsuperscript{2}\par}\vspace{0.2em}{\Affilfont\textsuperscript{1}Hong Kong University of Science and Technology\par}{\Affilfont\textsuperscript{2}The Chinese University of Hong Kong, Shenzhen\par}}
\makeatother
\reportnumber{}

\definecolor{darkblue}{rgb}{0.0,0.0,0.6}
\definecolor{darkred}{rgb}{0.7,0.0,0.0}
\hypersetup{
  pdffitwindow=true,pdfstartview={FitH},pdfnewwindow=true,
  colorlinks=true,linktocpage=true,
  linkcolor=darkred,urlcolor=darkblue,citecolor=darkblue,
  pdftitle={VA-TDM: A Unified View for Improving Few-Step Diffusion Models by Velocity Adversarial Distillation},
  pdfauthor={Yihong Luo, Tianyu Pang, Tianyang Hu}
}

\renewcommand{\titlefigure}{\noindent
\begin{minipage}{\linewidth}
    \centering
    \captionsetup{font=small,skip=3pt,hypcap=false}
    \includegraphics[width=1.0\linewidth]{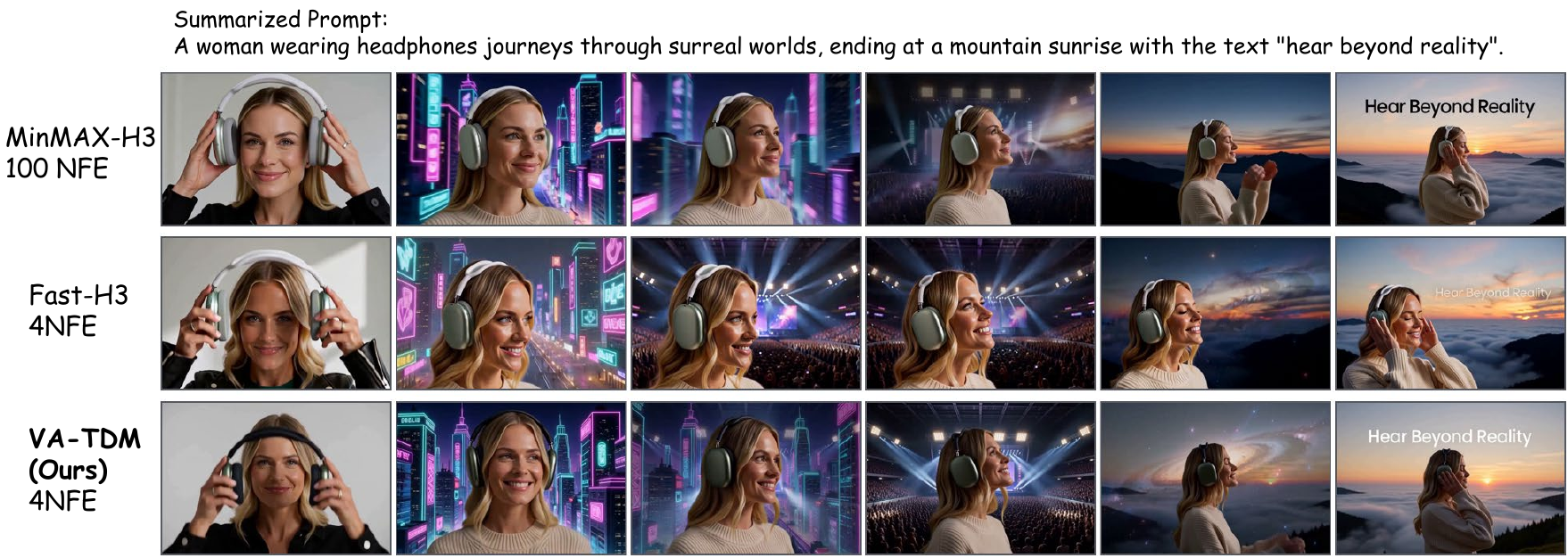}
    \captionof{figure}{Comparison of VA-TDM-H3 and competing baselines. Our VA-TDM delivers better visual quality and suffers less oversaturation than the few-step competitor. More samples in \cref{app:additional_visual_samples}.}
    \label{fig:teaser}
\end{minipage}
\par\vspace{10pt}
}

\begin{abstract}

Few-step distillation offers efficient generation with diffusion and flow models, but preserving quality with only a few sampling steps remains challenging. Distribution-matching methods via score distillation provide a strong foundation, yet they rely on accurately tracking the student’s changing distribution.
At intermediate generation steps in few-step sampling, student-score estimation can be difficult, and commonly used estimators may describe a different distribution from the one used to update the student.
Flow matching allows a more flexible approach to supervision through velocity fields constructed from the student’s sampling process. Building on this perspective, we introduce Velocity Adversarial Distillation (VAD), which uses a learned critic to compare student and teacher fields and turn their disagreement into feedback for learning.
VAD unifies score distillation, Fisher-based matching, and moment matching, also explaining existing heuristic MMD updates and delivering new training objectives.
We further develop VA-TDM, which follows trajectory distribution matching: providing supervision for partial denoising along the student trajectory. Beyond this, VA-TDM combines transition-aware online velocity learning and dual-form VAD supervision across intermediate student transitions.
Extensive results show that VA-TDM is powerful for learning few-step generators, achieving state-of-the-art four-step performance in both text-to-image and text-to-video generation across various backbones and scaling effectively to 33B MiniMax-H3. In particular, our VA-TDM-H3 notably outperforms the few-step Fast-H3 and is closer to the original many-step H3 according to human preference. Code is available at \url{https://github.com/Luo-Yihong/VA-TDM}.
\end{abstract}

\begin{document}

\maketitle

\clearpage
\begingroup
\setlength{\parskip}{5pt plus 1pt minus 1pt}
\section{Introduction}

Few-step distillation seeks to make diffusion and flow models efficient without sacrificing generation quality~\citep{ho2020denoising,sd3}. A central challenge is to provide effective supervision when a small number of student transitions must replace a long teacher trajectory. Trajectory distillation learns from individual teacher trajectories~\citep{song2023consistency}, whereas distribution matching allows the student to learn its own mapping by aligning student and teacher distributions~\citep{luo2023diff,yin2023one}. Trajectory Distribution Matching (TDM)~\citep{luo2025learning} extends this principle to intermediate transitions. Each step then learns a partial denoising task, distributing supervision across the student trajectory and providing a natural foundation for few-step distillation.

For intermediate supervision to be effective, however, the training signal must faithfully reflect the distribution produced at each student step. In score-based methods, generator updates depend on the difference between teacher and student scores, requiring an auxiliary model to track the evolving student distribution. Intermediate transitions complicate this estimation: the student's clean prediction and retained residual can be correlated, so perturbing its intermediate output generally differs from adding fresh Gaussian noise to the clean prediction.
Our analysis of TDM shows why this distinction matters. Direct score estimation for the perturbed intermediate output can involve high-variance targets, while clean-target denoising can estimate a different distribution from the one used for generator supervision.

These observations motivate supervision that explicitly accounts for the student's sampling process while allowing alternatives to its noisy-marginal score. Velocity targets offer this flexibility because they can be defined through paths with coupled endpoints~\citep{liu2022flow}, including dependencies inherited from a student transition. We build on this flexibility to introduce \emph{Velocity Adversarial Distillation} (VAD), a framework for constructing student velocity fields and using their discrepancies from the teacher to guide distillation.
Two prototypical velocity constructions are derived within VAD. \emph{A-form} is a rectified velocity that captures the expected clean prediction conditioned on a perturbed sample; \emph{B-form} is a score-equivalent Gaussian velocity for the student's perturbed distribution. The two forms thus express conditional-moment and distributional supervision within the same sampling process. A vector-valued critic estimates the discrepancy between the chosen student field and the teacher field, and the generator learns from this discrepancy.

The resulting framework provides a unified perspective along two axes: \emph{which field to match} and \emph{how to use its discrepancy to update the generator}. With different generator updates, A-form connects to conditional moment matching, while B-form connects reverse-KL score distillation and Fisher-based field matching. These connections also clarify the relationship between an intended objective and its practical update. In particular, VAD explains the heuristic update of MMD~\citep{mmd} and derives an alternative surrogate that recovers the exact gradient of MMD's desired objective.

\begin{wrapfigure}{r}{0.45\textwidth}
    \vspace{-8pt}
    \setlength{\parskip}{0pt}
    \centering
    \includegraphics[width=\linewidth]{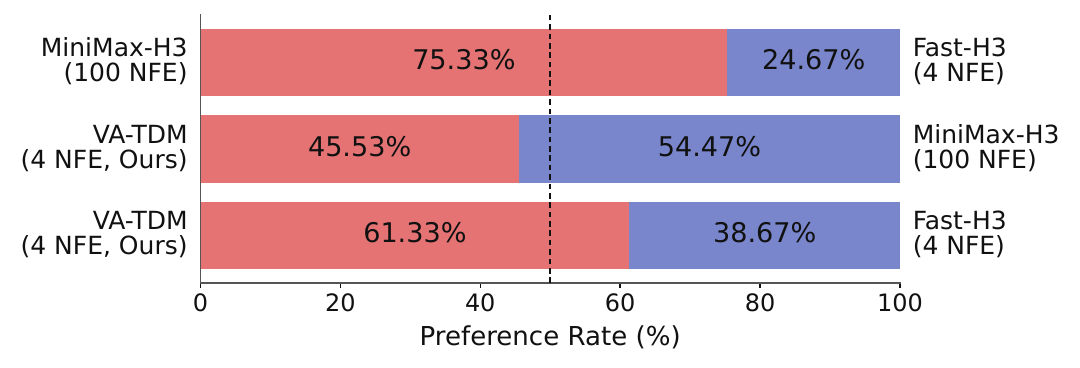}
    \par\vspace{-6pt}
    \setlength{\abovecaptionskip}{2pt}
    \setlength{\belowcaptionskip}{0pt}
    \caption{\small Human preference on video synthesis.}
    \label{fig:h3_user_study}
    \vspace{-6pt}
\end{wrapfigure}
To use both types of supervision along a few-step trajectory, we develop \emph{VA-TDM}, retaining TDM's supervision of intermediate transitions. VA-TDM combines A-form and B-form through a shared auxiliary backbone and blends complementary generator updates. Because samples from different student steps can follow different distributions even at the same noise level, we also condition field estimation on the transition being supervised.
This provides different and accurate supervision signals among different steps.

Overall, our unified VAD framework connects existing distillation updates and
derives new objectives, which VA-TDM combines with complementary velocity
fields and transition-aware estimation for four-step generation, achieving state-of-the-art performance across backbones and tasks.
On SD3.5-Medium, VA-TDM outperforms the compared few-step baselines across
prompt following, image quality, and preference metrics, improving
GenEval from TDM's 0.61 to 0.67 while increasing sample diversity.
On FLUX.2-Klein-Base 4B, it surpasses the released four-step model,
raising GenEval from 0.78 to 0.80 and OCR similarity from 0.64 to 0.69,
with diversity measured by LPIPS improving from 0.47 to 0.56.
Most notably, VA-TDM scales to MiniMax-H3~\citep{minimax2026h3},
a state-of-the-art open-source audio-video model with
33B parameters. In our user study (\cref{fig:h3_user_study}),
VA-TDM wins 61.33\% of comparisons against Fast-H3 at the same 4-NFE
budget and achieves a 45.53\% preference rate against the 100-NFE
teacher, compared with 24.67\% for Fast-H3.
These results demonstrate that VA-TDM substantially narrows the
perceptual quality gap to the teacher while requiring only
4-NFE.

\section{Preliminaries}

\spara{Diffusion and Flow Models}
Diffusion models~\citep{sohl2015deep,ho2020denoising} use the Gaussian
forward process $\bx_\tau=\alpha_\tau\bx_0+\sigma_\tau\beps$, where
$\bx_0\sim q_0$, $\beps\sim\mathcal N(\mathbf0,\mathbf I)$ independently,
and $q_\tau$ denotes the perturbed marginal. Time $\tau\in[0,1]$ runs
from clean data to the noise prior; conditioning on $\mathbf c$ is omitted.
For $\sigma_\tau>0$, denoising score matching (DSM) learns the marginal
score by regressing onto the conditional target:
\begin{equation}
\label{eq:dsm}
\bs_\tau^q(\bx_\tau)\coloneqq\nabla_{\bx_\tau}\log q_\tau(\bx_\tau)
= \mathbb E_{q_0(\rvx_0|\rvx_\tau=\rvx)}\!\left[ \nabla_{\bx_\tau}\log q_\tau(\rvx|\bx_0)  \right].
\end{equation}
Rectified flow~\citep{liu2022flow} uses the linear schedule
$\alpha_\tau=1-\tau$, $\sigma_\tau=\tau$ and velocity
$\bu_\tau(\bx)=\mathbb E[\beps-\bx_0\mid\bx_\tau=\bx]$.
For Gaussian forward processes, it is equivalent to diffusion models.

\spara{Trajectory Distribution Matching (TDM)}
TDM~\citep{luo2025learning} matches intermediate student and teacher
distributions along a $K$-step student trajectory $0=t_0<\cdots<t_K=1$.
For a transition $t=t_k\to s=t_{k-1}$, perturbing $\bx_s\sim p_{\theta,k}(\rvx_s)$
with $q(\bx_\tau\mid\bx_s)$, gives
student marginal $p_{\theta,\tau}^{\mathrm{path}}$.
Learning is performed by reverse-KL divergence~\citep{luo2023diff} along the student trajectory:
\begin{equation}
\label{eq:tdm_grad}
\nabla_\theta\mathcal L_{\mathrm{TDM}}
=\mathbb E_{k,t>\tau>s}\mathbb E_{q(\bx_\tau|\bx_s)p_{\theta,k}(\bx_s)}\!\left[
w_{k,\tau}\left(\frac{\partial\bx_\tau}{\partial\theta}\right)^{\!\top}
\left(\nabla_{\bx_\tau}\log p_{\theta,\tau}^{\mathrm{path}}(\bx_\tau)
-\bs_\tau^q(\bx_\tau)\right)\right].
\end{equation}
Teacher scores are supplied by the pretrained
model; student scores require online estimation.

\vspace{-2mm}
\section{Method}
\vspace{-1mm}
\label{sec:vad_method}

\spara{Problem Setup}
We consider distilling a pretrained conditional diffusion or flow model into
an efficient few-step generator. Let $q_t(\bx_t\mid\mathbf c)$ and
$\bu_t(\bx_t,\mathbf c)$ denote the teacher's marginal distribution and
velocity field at time $t\in[0,1]$, respectively, where $t=0$ corresponds to
clean samples and $t=1$ to the noise prior. Given a condition $\mathbf c$ and
a desired number of sampling steps $K$, our goal is to learn a student that
transforms noise into a sample through $K$ transitions while matching the
teacher's marginal distributions along the generation trajectory.

\spara{Overview}
We introduce \emph{Velocity Adversarial Distillation} (VAD) as a unified framework for distilling diffusion models through aligning student and teacher velocity fields by an adversarial game.
\cref{sec:vad_targets} motivates this formulation by examining
the challenges of student score estimation and introduces two
constructions of student velocity.
\cref{sec:vad_adversarial} develops the adversarial velocity
game, characterizes its matching equilibrium, and establishes
connections to existing distillation methods.
\cref{sec:vad_trajectory} presents practical generator and
critic updates and applies the framework along the student
trajectory, with transition-aware field estimation.

\vspace{-2mm}
\subsection{From Score Distillation to Velocity Distillation}
\vspace{-1mm}
\label{sec:vad_targets}

We focus on learning a one-step marginal transition from $\bx_t$ to a less noisy $\bx_s$, where
$0\leq s<t\leq1$, whose output distribution $p_{\theta,s}$ should match
the teacher marginal $q_s$.
Matching an intermediate distribution requires only partial denoising at each step, easing the learning task compared with requiring a clean output at every step, and is more efficient and effective~\citep{luo2025learning}.

Under the rectified-flow schedule, the student predicts a clean anchor and its residual given $\bx_t$:
\begin{equation}
\label{eq:vad_clean_prediction}
\hat\bx_0=g_\theta(\bx_t,t) = \bx_t - tv_\theta(\bx_t,t),\qquad
\hat\beps=\frac{\bx_t-(1-t)\hat\bx_0}{t}.
\end{equation}
Given $\hat\bx_0$ and $\hat\beps$, we can parameterize a DDIM-style intermediate transition~\citep{ddim}:
\begin{equation}
\label{eq:vad_transition}
\begin{aligned}
    \bx_s=(1-s)\hat\bx_0
+\eta s\,\hat\beps+s\sqrt{1-\eta^2}\mathbf z_s \sim p_\theta(\bx_s|\bx_t)
    \triangleq \mathcal{N}((1-s)\hat\bx_0
+\eta s\,\hat\beps, s^2(1-\eta^2) \mathbf I),
\end{aligned}
\end{equation}
where $0\leq\eta\leq1$ and $\mathbf z_s\sim\mathcal N(\mathbf0,\mathbf I)$. Note that the clean anchor is a parameterization of the transition; the
desired matching objective is imposed on $\bx_s$, not separately on $\hat\bx_0$.
By default, we let $\eta=1$, which forms a deterministic transition, while its
outputs remain random across inputs.

\spara{Distributional Supervision at a Perturbed Time}
To supervise the intermediate distribution, existing works in the line of score distillation~\citep{luo2023diff,luo2025learning} further perturb $\bx_s$
to $\bx_\tau$, where $\tau>s$. This smooths the distributions and makes the
teacher's knowledge at additional noise levels available for supervision.
For $s<\tau<1$, we use the teacher-compatible kernel
\begin{equation}
\label{eq:vad_kernel}
\begin{aligned}
q(\bx_\tau\mid\bx_s)
=\mathcal N(\bx_\tau;\alpha_{\tau\mid s}\bx_s,\beta_{\tau\mid s}^2\mathbf I), \quad
\alpha_{\tau\mid s}=\frac{1-\tau}{1-s},\qquad
\beta_{\tau\mid s}^2=\tau^2-\alpha_{\tau\mid s}^{\,2}s^2.
\end{aligned}
\end{equation}
Thus $\bx_\tau=\alpha_{\tau\mid s}\bx_s+
\beta_{\tau\mid s}\mathbf z_\tau$, where $\mathbf z_\tau\sim\mathcal{N}(\mathbf 0, \mathbf I)$.
The full sampling chain
factors into $q(\bx_s\mid\hat\bx_0,\hat\beps,\eta)$ and
$q(\bx_\tau\mid\bx_s)$.
Unless otherwise specified, expectations are
 taken over sampled $\tau$, this student transition and subsequent perturbation, with conditioning fixed.
Let $p_{\theta,s}$ denote the distribution of the student's
intermediate output $\bx_s$. After applying the perturbation
kernel, the resulting perturbed marginal distribution is
\begin{equation}
\label{eq:vad_perturbed_marginal}
p_{\theta,\tau}^{\mathrm{path}}(\bx_\tau)
=
\int q(\bx_\tau\mid\bx_s)\,
p_{\theta,s}(\bx_s)\,\mathrm d\bx_s.
\end{equation}
This is the student distribution whose score is required for
reverse-KL supervision at time $\tau$. Because $\bx_s$ retains
the residual inherited from the student transition, this
distribution generally differs from one obtained by adding
independent Gaussian noise directly to $\hat\bx_0$.

If the optimization is formed by reverse-KL following previous works~\citep{luo2025learning}, we would require the score difference
$\nabla\log p_{\theta,\tau}^{\mathrm{path}}-\nabla\log q_\tau$. The teacher supplies the latter score, whereas the former must be learned online for the current student's actual perturbation distribution.

\spara{The Difficulty of Learning the Student Score} DSM can estimate this score $\nabla\log p_{\theta,\tau}^{\mathrm{path}}$ by regressing onto $\nabla_{\bx_\tau}\log q(\bx_\tau\mid\bx_s)$.
\textit{Although population-correct, under the reference Gaussian forward process this target has larger conditional variance than standard DSM}, which regresses $\nabla_{\bx_\tau}\log q(\bx_\tau\mid\bx_0)$~\citep{rl_awm}, thereby making the online score harder to learn.

\begin{remark}[Mismatch in TDM]
To avoid this difficulty, TDM learns its online score through
clean-target denoising with importance
sampling~\citep{luo2025learning}. Although importance
sampling allows training samples to be drawn from
$p_{\theta,\tau}^{\mathrm{path}}$, the objective still estimates
the score of the distribution obtained by independently perturbing
$\hat\bx_0$ to $\tau$. This generally differs from the score of
$p_{\theta,\tau}^{\mathrm{path}}$ required by the generator update,
introducing a distribution mismatch
(See detailed derivation in \cref{app:vad_score_mismatch}).
\end{remark}

\spara{Velocity as a Choice of Path and Coupling}
This distinction motivates a shift from score distillation to velocity distillation, which can define student fields directly from the actual sampling chain. In particular, \textit{\textbf{Velocity targets can be induced by differentiable
paths with correlated or non-Gaussian endpoint couplings}}, therefore giving us greater flexibility.
In the student transition $t\to s$, $\hat\bx_0$ and $\hat\beps$ share
the same input and can be dependent, alongside optional fresh noise; the
forward perturbation $s\to\tau$ adds independent Gaussian noise. This gives
two choices: use the clean-to-perturbed coupling induced by the composite
process $t\to s\to\tau$, or follow the $s\to\tau$ path alone.
Throughout the remaining sections, $p_{\theta,\tau}$ abbreviates
$p_{\theta,\tau}^{\mathrm{path}}$, the perturbed marginal.

\spara{A-form: Rectified Velocity}
The composite process $t\to s\to\tau$ couples the clean prediction $\hat\bx_0$
to $\bx_\tau$. Expanding \cref{eq:vad_transition} yields
\begin{equation}
\label{eq:vad_composite_noise}
\begin{aligned}
\bx_\tau&=(1-\tau)\hat\bx_0+\boldsymbol\xi_{\theta,s,\tau},\\
\boldsymbol\xi_{\theta,s,\tau}
&=\alpha_{\tau\mid s}\eta s\,\hat\beps
+\alpha_{\tau\mid s}s\sqrt{1-\eta^2}\mathbf z_s
+\beta_{\tau\mid s}\mathbf z_\tau.
\end{aligned}
\end{equation}
The effective noise $\beps_{\mathrm{eff}}=\boldsymbol\xi_{\theta,s,\tau}/\tau$
contains the residual inherited from the student and standard Gaussian noise.
This can deliver a corresponding rectified velocity coupled with the endpoint:
\begin{equation}
\label{eq:vad_b_target}
\bv_\tau^A(\bx_\tau\mid\bx_s,\hat\bx_0)
=\beps_{\mathrm{eff}}-\hat\bx_0
=\frac{\bx_\tau-\hat\bx_0}{\tau}.
\end{equation}
The A-form rectified velocity uses the clean prediction as its anchor through the
composite coupling. It produces reduced training variance when $s>0$, which benefits learning of the online field. This marginal velocity is an affine reparameterization of a moment $\mathbb E[\hat\bx_0\mid\bx_\tau]$.

\spara{B-form: Gaussian Velocity}
The perturbation $s\to\tau$ defines a path starting at $\bx_s$.
Differentiating
this path with respect to $\tau$,
and using
$\mathbf z_\tau=(\bx_\tau-\alpha_{\tau\mid s}\bx_s)/\beta_{\tau\mid s}$,
gives the conditional velocity
\begin{equation}
\label{eq:vad_a_target}
\bv_\tau^B(\bx_\tau\mid\bx_s,\hat\bx_0)
=\dot\alpha_{\tau\mid s}\bx_s
+\frac{\dot\beta_{\tau\mid s}}{\beta_{\tau\mid s}}
\bigl(\bx_\tau-\alpha_{\tau\mid s}\bx_s\bigr).
\end{equation}
Dots denote $\tau$ derivatives with $s$ fixed.
Its conditional average is the marginal velocity of the
$s\to\tau$ perturbation path. This marginal velocity is an affine reparameterization of the
score of the actual perturbed distribution (see proof in \cref{app:vad_unified_connections_proof}).

\spara{The Induced Student Field}
We use the common notation $\bv_\tau^F(\bx_\tau\mid\bx_s,\hat\bx_0)$
for these two conditional targets, with $F\in\{A,B\}$. The field to be compared with the teacher is
\begin{equation}
\label{eq:vad_mean_field}
\bar\bv^F_{\theta,\tau}(\bx)
=\mathbb E_{p_\theta(\bx_s,\hat\bx_0\mid \bx_\tau=\bx)}\!\left[
\bv_\tau^F(\bx\mid\bx_s,\hat\bx_0)\right].
\end{equation}
Both formulations can be computed via student samples, but their marginal fields require averaging over the implicit student distribution.

\vspace{-2mm}
\subsection{Velocity Adversarial Distillation}
\vspace{-1mm}
\label{sec:vad_adversarial}
\label{sec:vad_connections}

Diff-Instruct has discussed that reverse-KL-based score distillation is related to GANs~\citep{luo2023diff,goodfellow2014generative}. We further find that, after shifting from score distillation to velocity distillation, explicitly defining an adversarial game that aims to align the teacher velocity with the student velocity forms a unified framework and view, which deepens the understanding of existing methods and might inspire the design of new distillation losses.

\spara{Matching Implicit Velocity}
We propose to align the student and teacher velocity fields through an adversarial game defined by their inner products with a vector-valued critic $\bff$, i.e., $\bff^\top\bv$. In particular, the critic maximizes its alignment with the teacher velocity and minimizes its alignment with the student velocity, subject to a quadratic penalty. The student opposes the critic by maximizing its own velocity's alignment with the critic, with the critic held fixed.

For brevity, we write
$\bv_\tau^F \equiv
\bv_\tau^F(\bx_\tau\mid\bx_s,\hat\bx_0)$
for the sampled conditional velocity, and let $\mathbb E$ denote
expectation over $\tau$ and the full sampling chain
$p_\theta(\bx_t,\bx_s,\hat\bx_0)q(\bx_\tau\mid\bx_s)$.
Since $\bff$ depends only on $\bx_\tau$, replacing the sampled velocity $\bv_\tau^F$
by its conditional-mean velocity $\bar \bv_\tau^F$ leaves the expected inner product unchanged:
\begin{equation}
\mathbb E\!\left[
\bff(\bx_\tau)^\top\bv_\tau^F
\right]
=
\mathbb E_{\bx_\tau\sim p_{\theta,\tau}}\!\left[
\bff(\bx_\tau)^\top\bar\bv_{\theta,\tau}^F(\bx_\tau)
\right].
\end{equation}
This identity motivates two generator losses with the same expected
value but different gradient conventions.
The \textit{pathwise update} differentiates through the full sampling
process, changing both the outer sampling distribution and the student velocity
at those locations. The \textit{conditional-mean update} stops the gradient of the sampling distribution and adjusts only the student velocity
field.

In particular, the adversarial game in the velocity space can be formulated as:
\begin{align}
\max_{\bff}\quad V_F(\bff,\theta)
&=\mathbb E\!\left[
\bff(\bx_\tau)^\top
\bigl(\bu_\tau(\bx_\tau)-\bar\bv_{\theta,\tau}^F\bigr)
-\tfrac12\|\bff(\bx_\tau)\|^2
\right],
\label{eq:vad_alternating_game}
\\
\min_\theta\quad
\mathcal L_{\mathrm{path}}^F(\theta;\bff)
&=-\mathbb E\!\left[
\bff(\bx_\tau)^\top\bv_\tau^F
\right],
\nonumber
\\
\text{or}\qquad
\min_\theta\quad
\mathcal G_{\mathrm{mean}}^F(\theta;\bff)
&=-\mathbb E_{\bx_\tau\sim p_{\operatorname{sg}(\theta),\tau}}
\!\left[
\bff(\bx_\tau)^\top
\bar\bv_{\theta,\tau}^F(\bx_\tau)
\right].
\label{eq:vad_mean_path_loss}
\end{align}
Here $\operatorname{sg}$ denotes stop-gradient. Only the outer
distribution is stopped in $\mathcal G_{\mathrm{mean}}^F$;
$\bar\bv_{\theta,\tau}^F$ retains its parameter dependence.
The following lemma reveals the formulation of the optimal critic:
\begin{lemma}[Optimal critic]
\label{lem:vad_optimal_witness}
For a fixed student $\theta$, under the regularity and critic assumptions
detailed in \cref{app:vad_equilibrium_assumptions}, the maximizing
critic is unique up to
$p_{\theta,\tau}$-null sets and satisfies
\begin{equation}
\label{eq:vad_optimal_witness}
\bff_F^\star(\bx)=\bu_\tau(\bx)-\bar\bv^F_{\theta,\tau}(\bx).
\end{equation}
\end{lemma}
The proof is given in \cref{app:vad_optimal_witness_proof}.

\spara{Matching Equilibrium}
The optimal critic in \cref{lem:vad_optimal_witness} is the velocity
residual. Its zero identifies the following equilibrium of the two-player game.
\begin{proposition}[Matching equilibrium]
\label{thm:vad_matching_equilibrium}
Under the regularity, critic, and realizability assumptions in
\cref{app:vad_equilibrium_assumptions}, the game in
\cref{eq:vad_alternating_game} admits a Nash equilibrium satisfying
\begin{equation}
\label{eq:vad_optimal_matching}
\bar\bv^F_{\theta^\star,\tau}=\bu_\tau
\quad p_{\theta^\star,\tau}\text{-almost everywhere},
\qquad \bff^\star\equiv\mathbf0.
\end{equation}
It attains zero velocity discrepancy, the minimum of the maximized critic payoff.
\end{proposition}
The proof is in \cref{app:vad_matching_equilibrium_proof}.

Directly computing the conditional-mean update is intractable because
it requires differentiating the implicit field
$\bar\bv_{\theta,\tau}^F$ with respect to $\theta$.
The following lemma gives a general flow-product update through
a tractable surrogate whose gradient agrees with the
conditional-mean update.

\begin{lemma}[General Flow-Product Update]
\label{lem:vad_flow_product}
Fix a reference parameter $\theta_0=\bar\theta$ and let
$\mathbf h_0^F(\bx)=\bar\bv_{\theta_0,\tau}^F(\bx)$.
Hold the parameters of $\bff$ and $\mathbf h_0^F$ fixed, while retaining
their input derivatives. Define
\begin{equation}
\label{eq:vad_flow_product_surrogate}
\begin{aligned}
\mathcal G_{\mathrm{fp}}^F(\theta;\bff,\mathbf h_0)
:={}-\mathbb E_{p_\theta(\bx_t,\bx_s,\hat\bx_0)q(\bx_\tau\mid\bx_s)}
\left[\bff(\bx_\tau)^\top
\left(\bv_\tau^F(\bx_\tau\mid\bx_s,\hat\bx_0)
-\mathbf h_0^F(\bx_\tau)\right)\right].
\end{aligned}
\end{equation}
Under the regularity conditions detailed in \cref{app:vad_flow_product_proof},
\begin{equation}
\label{eq:vad_flow_product_gradient}
\begin{aligned}
\left.\nabla_\theta\mathcal G_{\mathrm{fp}}^F\right|_{\theta_0}
&=\left.\nabla_\theta\mathcal G_{\mathrm{mean}}^F
(\theta;\theta_0,\bff)\right|_{\theta_0}
=-\mathbb E_{\bx_\tau\sim p_{\theta_0,\tau}}
\left[\bff(\bx_\tau)^\top
\left.\partial_\theta\bar\bv_{\theta,\tau}^F(\bx_\tau)\right|_{\theta_0}\right].
\end{aligned}
\end{equation}
\end{lemma}
The proof is given in \cref{app:vad_flow_product_proof}.

\begin{theorem}[Matching meanings and algorithmic unification]
\label{thm:vad_unified_connections}
Use the compatible Gaussian kernel in \cref{eq:vad_kernel} and an exact
teacher. For the gradient identities, fix the input law and times
$0\leq s<t\leq1$, $s<\tau<1$, take $\eta=1$, and assume exact critics and
reference fields with the differentiability conditions of
\cref{app:vad_flow_product_proof}.

\textup{(i) Matching.}
A-form velocity matching is equivalent to the conditional-moment matching
\begin{equation}
\label{eq:vad_b_matching}
\mathbf m_\theta(\bx)
\coloneqq\mathbb E[\hat\bx_0\mid\bx_\tau=\bx]
=\mathbf m_q(\bx)
\coloneqq\bx-\tau\bu_\tau(\bx).
\end{equation}
B-form velocity matching is equivalent to distribution matching
$p_{\theta,\tau}=q_\tau$, and hence $p_{\theta,s}=q_s$.

\textup{(ii) Equivalence at $s=0$.}
At $s=0$, A and B coincide, with identical matching conditions and population gradients under the same generator update.

\textup{(iii) Unifying Algorithms.}
The two forms and two gradient choices yield the correspondences below,
up to positive time weights.
\begin{center}
\small
\begin{tabular}{@{}lp{0.43\linewidth}p{0.43\linewidth}@{}}
\toprule
Form & Pathwise w/ detached critic $\mathrm{sg}(\bff(\rvx_\tau))$ & Flow-product w/ non-detach critic $\bff(\rvx_\tau)$ \\
\midrule
A & MMD practical update~\citep{mmd} & Exact Moment Matching \\
B & Reverse-KL~\citep{luo2023diff,luo2025learning,yin2023one}
  & Fisher Divergence~\citep{luo2024onestep,sid} \\
\bottomrule
\end{tabular}
\end{center}
\end{theorem}

The proof and time coefficients are in
\cref{app:vad_unified_connections_proof}.

\spara{MMD: Objective and Approximate Update}
MMD~\citep{mmd}
aims to minimize a squared conditional-moment discrepancy to align the teacher moment and student moment. Because
differentiating the student's implicit moment is intractable,
it adopts a heuristic approximate gradient to update the student generator.
\textbf{\textit{Our unified framework provides both an exact gradient and a mechanistic explanation
of this heuristic}}: A-form flow-product updates recover the intended
objective's exact gradient under MMD's fixed-outer-measure convention,
while A-form pathwise updates with a detached critic recover the surrogate
update as a generator update in the adversarial velocity game, up to
positive time weighting on the same sampling chain.
\cref{app:vad_mmd_connection} explicitly gives the target objective,
the approximation leading to the surrogate, and both gradient identities.

\begin{figure}[!t]
    \centering
    \includegraphics[width=1.0\linewidth]{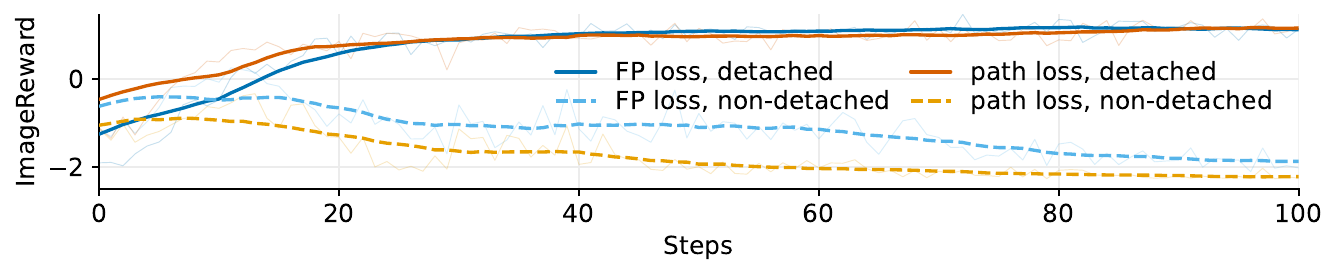}
    \vspace{-8mm}
    \caption{ Comparison between detached critic and non-detached critic. We found that the non-detached critic leads to unstable training and collapse.}
    \label{fig:detach_compare}
\end{figure}

\vspace{-2mm}
\subsection{VA-TDM: Distillation Along the Student Trajectory}
\vspace{-1mm}
\label{sec:vad_trajectory}

\spara{Generator Learning}
We observe unstable training when gradients are propagated through the
critic output (\cref{fig:detach_compare}). We therefore stop the entire critic output in the practical
update. Let $\operatorname{sg}$ denote this stop-gradient operation and define
\begin{equation}
\label{eq:vad_stopped_updates}
\begin{aligned}
\mathcal L_{\mathrm{path,sg}}^F
&=-\mathbb E_{p_\theta(\bx_t,\bx_s,\hat\bx_0)q(\bx_\tau\mid\bx_s)}
\left[\operatorname{sg}(\bff^F(\bx_\tau))^\top
\bv_\tau^F(\bx_\tau\mid\bx_s,\hat\bx_0)\right],\\
\mathcal L_{\mathrm{fp,sg}}^F
&=-\mathbb E_{p_\theta(\bx_t,\bx_s,\hat\bx_0)q(\bx_\tau\mid\bx_s)}
\left[\operatorname{sg}(\bff^F(\bx_\tau))^\top
\left(\bv_\tau^F(\bx_\tau\mid\bx_s,\hat\bx_0)
- \bar{\bv}_{\theta,\tau}^F\right)\right].
\end{aligned}
\end{equation}
Note that $L_{\mathrm{fp,sg}}^F$ and $L_{\mathrm{path,sg}}^F$ correspond to different adversarial flexibility for the generator: the latter allows the generator to cheat the critic by altering both the outer sampling distribution and the velocity. Since the balance of the adversarial game is crucial for performance, we propose optimizing the hybrid loss to provide an explicit knob for controlling the balance.
\begin{equation}
\label{eq:vad_mixed_loss}
\begin{aligned}
\mathcal L_{\mathrm{mix}}^F
&=(1-\lambda)\mathcal L_{\mathrm{path,sg}}^F
+\lambda\mathcal L_{\mathrm{fp,sg}}^F.
\end{aligned}
\end{equation}

\spara{Critic Learning}
With the help of \cref{lem:vad_optimal_witness}, we parameterize the critic as
$\bff_\psi^F=\bu_\tau-\mathbf h_\psi^F$ and fit the student field with
\begin{equation}
\label{eq:vad_field_training}
\mathcal L_{\mathrm{field}}^F(\psi)
=\mathbb E_{p_\theta(\bx_s,\hat\bx_0)q(\bx_\tau\mid\bx_s)}
\left[\|\mathbf h_\psi^F(\bx_\tau)
-\bv_\tau^F(\bx_\tau\mid\bx_s,\hat\bx_0)\|^2\right].
\end{equation}
To retain the pre-trained knowledge, we add the regularization term
$\lambda_r\mathbb E[\|\mathbf h_\psi^F(\bx_\tau)
-\bu_\tau(\bx_\tau)\|_2^2]$, with $\lambda_r>0$,
under the same sampling distribution.
The optimal minimizer then becomes
\[
\mathbf h_{\psi^\star}^F(\bx)
=\frac{\bar\bv_{\theta,\tau}^F(\bx)
+\lambda_r\bu_\tau(\bx)}{1+\lambda_r}.
\]
We therefore should correctly parameterize the critic as
$\bff_\psi^F=(1+\lambda_r)(\bu_\tau-\mathbf h_\psi^F)$
to perform best.

\spara{Dual-Form Velocity Game with a Shared Backbone}
The A-form produces low-variance online field learning, but it only captures the conditional-mean moment of the distribution; the B-form induces high-variance online field learning, yet it encodes the full distributional information. Since no single velocity construction enjoys both benefits, we propose to optimize the adversarial games induced by the two velocity constructions simultaneously, the generator loss becomes:
\begin{equation}
\mathcal L_{\mathrm{mix}} = \lambda_m \mathcal L_{\mathrm{mix}}^{A} + (1-\lambda_m) \mathcal L_{\mathrm{mix}}^{B},
\label{eq:vad_dual_form_loss}
\end{equation}
where each term follows \cref{eq:vad_mixed_loss} with its corresponding velocity target. To avoid the cost of maintaining two online fields, we propose a shared-backbone strategy: the two field networks share the backbone with separate output heads. \cref{tab:ablation} shows that this sharing strategy not only reduces memory and training overhead, but also improves performance.

\spara{Distillation Along the Student Trajectory}
Matching only the final output places the entire distributional correction
at the end of generation. We have discussed the algorithm for learning one-step marginal transition from $\rvx_t$ to $\rvx_s$. Following TDM~\citep{luo2025learning}, we instead
apply VAD to intermediate student transitions along the student trajectory. Let
$0=t_0<t_1<\cdots<t_K=1$ be a sampling grid. Starting from
$\bx_{t_K}\sim\mathcal N(\mathbf0,\mathbf I)$, the student repeatedly applies
the student transition in \cref{eq:vad_transition} with
$(t,s)=(t_k,t_{k-1})$ and $\eta=1$.
Each transition learns a distributional mapping from its current noisy
state to the next state; no pointwise teacher ODE solution is needed.

\spara{On-Trajectory ODE Regression Warm-up}
We initialize VA-TDM with ODE regression along the student's own
sampling trajectory. For a sampled transition $(t_k,t_{k-1})$, we
roll out the student from noise to obtain $\bx_{t_k}$ and integrate
the teacher ODE over the same interval using multiple solver steps:
\begin{equation}
    \bx^{\mathrm{tea}}_{t_{k-1}}
    =
    \operatorname{ODESolver}
    (\bx_{t_k}, v_{\mathrm{tea}}, t_k, t_{k-1}).
\end{equation}
We train the student transition in \cref{eq:vad_transition},
whose output is denoted by $\hat{\bx}_{t_{k-1}}$, with
\begin{equation}
    \mathcal{L}_{\mathrm{warm}}
    =
    \mathbb{E}_{k,\,\bx_{t_k}\sim p_{\theta,t_k}}
    \left[
        d\left(
            \hat{\bx}_{t_{k-1}},
            \operatorname{sg}\!\left[
                \bx^{\mathrm{tea}}_{t_{k-1}}
            \right]
        \right)
    \right],
\end{equation}
where $d(\cdot,\cdot)$ is the distance metric and $p_{\theta,t_k}$ is the student marginal at $t_k$ and
$\operatorname{sg}$ denotes stop-gradient.
Following the connection between ODE regression and distribution
matching in TDM~\citep{luo2025learning}, perfect pointwise matching
also implies trajectory distribution alignment.
Specifically, assuming exact teacher integration and that every
transition is sampled with positive probability,
\begin{equation}
    \mathcal{L}_{\mathrm{warm}}=0,
    \qquad
    p_{\theta,t_K}=p_{\mathrm{tea},t_K}
    \quad\Longrightarrow\quad
    p_{\theta,t_k}=p_{\mathrm{tea},t_k},
    \quad k=0,\ldots,K,
\end{equation}
where $p_{\mathrm{tea},t_k}$ denotes the teacher ODE marginal.
This follows by induction: each perfectly matched transition
pushes the shared current marginal to the same next marginal.
Thus, ODE regression imposes a stronger pointwise constraint
whose exact satisfaction also fulfills trajectory distribution
matching. Therefore, this initialization directly targets the intermediate transitions and marginals optimized by VA-TDM. We subsequently switch to VAD, relaxing
pointwise regression to distributional alignment.

\spara{Transition-Aware Field Estimation}
In both our framework and existing score distillation work, it is crucial for the auxiliary score to accurately track the student distribution in order to achieve strong performance. However,\textit{we identify an aspect that has been consistently overlooked in prior work~\citep{dmd2,ddmd}}: in the few-step generation setting, student transitions with different step indices generally induce distinct noisy distributions, yet existing methods typically model them with a shared auxiliary score, which theoretically precludes the possibility of precisely tracking the student distribution.
TDM tries to avoid this by assigning student transitions with different step indices to disjoint perturbation intervals, which is partially effective but sacrifices part of supervision signals.
To resolve this issue, \textit{\textbf{we propose that the field network should additionally receive the identity of the transition}}—for instance, in our setting, the target timestep of the student transition can be injected as extra input, i.e., learn $h_\psi(\bx_\tau,\tau,s)$ as online velocity. Results in \cref{tab:ablation} show that this modification is essential for providing accurate supervision signals, and substantially improves diversity and fine-grained rendering of images, such as text rendering.

\vspace{-1.5mm}
\section{Experiments}
\vspace{-0.5mm}
\suppressfloats[t]
\label{sec:vatdm_experiments}

\spara{Models and Baselines}
We distill SD3.5-Medium~\citep{sd3}, FLUX.2-Klein-Base 4B~\citep{flux2klein} and MiniMax-H3~\citep{minimax2026h3}
into four-step generators.
For SD3.5-Medium, the baselines include
DMD2~\citep{dmd2}, D-DMD~\citep{ddmd}, DPDMD~\citep{dpdmd}, and TDM~\citep{luo2025learning}.
For FLUX.2 4B, we compare with the officially released four-step model.

\spara{Evaluation Protocol}
We evaluate images along four complementary dimensions. \textit{Prompt following} is assessed using GenEval~\citep{ghosh2023geneval}
for compositional generation and OCR for visual text rendering.
\textit{Image quality} is measured by Aesthetic
Score~\citep{schuhmann2022laion}, while \textit{preference} is assessed with
ImageReward~\citep{xu2023imagereward}, PickScore~\citep{kirstain2023pick},
HPSv2, and HPSv3; these quality and preference metrics are evaluated on
DrawBench~\citep{drawbench}.
Finally, \textit{diversity} is measured by
average pairwise LPIPS~\citep{zhang2018perceptual} between independently
generated images by the same prompt.
For audio-video generation, we conduct anonymized pairwise human
comparisons of audio-visual quality and prompt alignment
(\cref{app:h3_user_study_details}).

\begin{table*}[!t]
\centering
\scriptsize
\renewcommand{\arraystretch}{1.05}
\caption{\small\textbf{Systematic evaluation.}
Best / second-best few-step results per backbone are
\textbf{bold} / \underline{underlined}.}
\label{tab:baselines}
\footnotesize
\setlength{\tabcolsep}{3pt}
\resizebox{\linewidth}{!}{\begin{tabular}{lccccccccc}
\toprule
\multirow{2}{*}{\textbf{Model}} & \multirow{2}{*}{\textbf{NFE}} & \multicolumn{2}{c}{\textbf{Verifiable}} & \textbf{ImgQ} & \multicolumn{4}{c}{\textbf{Preference}} & \textbf{Div.} \\
\cmidrule(lr){3-4}\cmidrule(lr){5-5}\cmidrule(lr){6-9}\cmidrule(l){10-10}
& & \textbf{GenEval} & \textbf{OCR} & \textbf{Aest.} & \textbf{ImgRwd} & \textbf{PickScore} & \textbf{HPSv2} & \textbf{HPSv3} & \textbf{LPIPS} \\
\midrule
SD3.5-Medium (teacher, CFG=4.5) & 50$\times$2 & 0.65 & 0.58 & 5.37 & 0.86 & 0.8628 & 0.2805 & 9.78 & 0.61 \\
SD3.5-Medium (teacher, CFG=7) & 50$\times$2 & 0.67 & 0.62 & 5.37 & 0.96 & 0.8640 & 0.2854 & 9.87 & 0.60 \\
\midrule
DMD2 & 4 & 0.60 & 0.53 & 5.55 & 1.00 & 0.8648 & \underline{0.2977} & 11.18 & 0.53 \\
D-DMD & 4 & 0.55 & 0.56 & 5.51 & 0.99 & 0.8567 & 0.2897 & 10.93 & 0.57 \\
DPDMD & 4 & \underline{0.62} & 0.33 & 5.28 & 0.80 & 0.8530 & 0.2707 & 8.13 & \textbf{0.59} \\
TDM & 4 & 0.61 & \underline{0.57} & \underline{5.58} & \underline{1.01} & \underline{0.8679} & 0.2969 & \underline{11.54} & 0.57 \\
\midrule
\textbf{VA-TDM (Ours)} & 4 & \textbf{0.67} & \textbf{0.59} & \textbf{5.65} & \textbf{1.02} & \textbf{0.8702} & \textbf{0.3010} & \textbf{11.89} & \underline{0.58} \\
\midrule
\midrule
FLUX.2-Klein-Base 4B (teacher, CFG=4.0) & 50$\times$2 & 0.77 & 0.60 & 5.31 & 0.85 & 0.8605 & 0.2768 & 10.28 & 0.56 \\
FLUX.2-Klein-Base 4B (teacher, CFG=7) & 50$\times$2 & 0.80 & 0.64 & 5.38 & 0.93 & 0.8620 & 0.2840 & 10.58 & 0.54 \\
\midrule
FLUX.2-Klein 4B & 4 & \underline{0.78} & \underline{0.64} & \underline{5.28} & \underline{1.04} & \textbf{0.8746} & \underline{0.2886} & \underline{11.27} & \underline{0.47} \\
\midrule
\textbf{VA-TDM (Ours)} & 4 & \textbf{0.80} & \textbf{0.69} & \textbf{5.43} & \textbf{1.07} & \underline{0.8694} & \textbf{0.2930} & \textbf{11.58} & \textbf{0.56} \\
\bottomrule
\end{tabular}}
\end{table*}

\vspace{-1mm}
\subsection{Main Results}
\vspace{-0.5mm}

\spara{Quantitative Results}
\cref{tab:baselines} compares VA-TDM with few-step baselines and their
teachers on text-to-image generation. At four NFE, VA-TDM improves compositional prompt following and
text rendering on both backbones. On SD3.5-Medium, it leads the compared
few-step models on all verifiable, quality, and preference metrics,
raising GenEval from the strongest baseline TDM's 0.61 to 0.67 while increasing TDM's diversity.
On FLUX2, gains in prompt following and most quality and preference metrics
are accompanied by an LPIPS increase from 0.47 to 0.56.
Together, the results demonstrate the state-of-the-art performance of our proposed VA-TDM in few-step generation.

\spara{Qualitative Results}
Across visual comparisons in image and video generation (\cref{fig:vatdm_visual_comparison,fig:teaser}), VA-TDM consistently delivers better visual quality compared with few-step baselines, significantly approaching the many-step base model. \textbf{\textit{Moreover, we find that previous methods either suffer from oversaturation or excessively dark outputs, whereas VA-TDM effectively avoids these undesirable artifacts.}}

\begin{figure}[!t]
    \centering
    \includegraphics[width=1.0\linewidth]{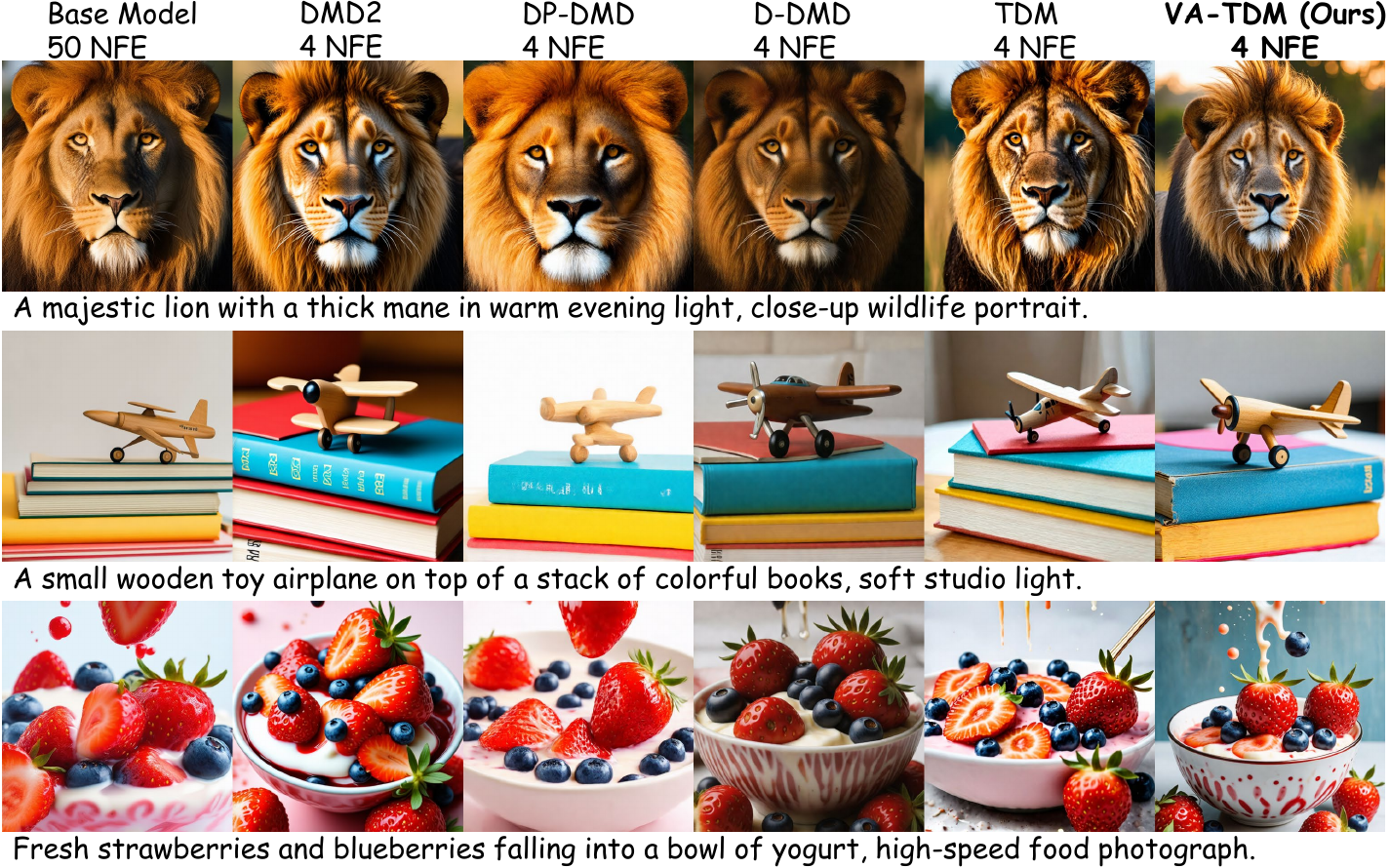}
    \vspace{-2.5mm}
    \caption{\textbf{Qualitative comparison} among VA-TDM and competing baselines.}
    \vspace{-2mm}
    \label{fig:vatdm_visual_comparison}
\end{figure}

\vspace{-2mm}
\subsection{Ablation Study}
\vspace{-2mm}

We examine the components introduced in \cref{sec:vad_trajectory} using four-step SD3.5-Medium
(\cref{tab:ablation}). Following \cref{eq:vad_dual_form_loss},
$\lambda_m$ mixes the A-form and B-form games; the separate coefficient
$\lambda$ in \cref{eq:vad_mixed_loss} controls the path/flow-product
mixture within each form, and $\lambda_r$ weights the online field's
regularization toward the teacher velocity. Starting from
$\lambda_m=0.8$, the final three rows report cumulative configurations
with generator-update mixing, field regularization, and critic correction.

\spara{Dual-Form Velocity Mixing}
The A-form rectified velocity and B-form Gaussian velocity in
\cref{sec:vad_targets} yield complementary results. A-form alone
improves GenEval over B-form alone, but has lower OCR and HPS scores.
Combining the two games produces higher OCR and
ImageReward than either endpoint while retaining the A-form diversity.
This supports combining their supervision.

\spara{Shared Backbone} Sharing the online-field backbone between the two forms not only reduces training cost but also improves performance. We believe that this is because the high-variance B-form is difficult to learn on its own, whereas the shared backbone allows the network to learn from both online fields jointly, which potentially leads to better internal representations and thus provides better supervision signals.

\spara{Path/Flow-Product Mixing}
The mixed update notably improves GenEval and diversity, while the other metrics remain roughly unchanged. This indicates that the mixed update allows us to find a better-performing sweet spot.

\spara{Online Field Regularization and Regularization-Corrected Critic} We augment the field regression in \cref{eq:vad_field_training} with
the teacher-velocity regularizer,
using $\lambda_r=1$. Even without critic correction, this also slightly improves
GenEval, OCR, supporting the benefit of retaining teacher information during
online field learning. By applying the regularization-corrected
critic $\bff_\psi^F=(1+\lambda_r)(\bu_\tau-\mathbf h_\psi^F)$ and adjusted loss coefficients, the
resulting configuration notably improves the performance and achieves the strongest verifiable, quality,
and preference scores among variants.

\begin{table*}[!t]
\centering
\scriptsize
\renewcommand{\arraystretch}{1.05}
\caption{\small\textbf{Ablation study.}
Best / second-best results are \textbf{bold} / \underline{underlined}.
}
\label{tab:ablation}
\footnotesize
\setlength{\tabcolsep}{3pt}
\resizebox{\linewidth}{!}{\begin{tabular}{lcccccccc}
\toprule
\multirow{2}{*}{\textbf{Configuration}} & \multicolumn{2}{c}{\textbf{Verifiable}} & \textbf{ImgQ} & \multicolumn{4}{c}{\textbf{Preference}} & \textbf{Div.} \\
\cmidrule(lr){2-3}\cmidrule(lr){4-4}\cmidrule(lr){5-8}\cmidrule(l){9-9}
& \textbf{GenEval} & \textbf{OCR} & \textbf{Aest.} & \textbf{ImgRwd} & \textbf{PickScore} & \textbf{HPSv2} & \textbf{HPSv3} & \textbf{LPIPS} \\
\midrule
$\lambda_m{=}1$ (A-form only) & 0.64 & 0.55 & 5.53 & 1.00 & 0.8671 & 0.2949 & 11.20 & \underline{0.58} \\
$\lambda_m{=}0$ (B-form only) & 0.61 & 0.57 & 5.57 & 0.99 & 0.8665 & \underline{0.2987} & \underline{11.47} & \textbf{0.59} \\
$\lambda_m{=}0$ w/o transition-aware field & 0.58 & 0.53 & 5.56 & 0.99 & 0.8655 & 0.2964 & 11.13 & 0.56 \\
\midrule
\midrule
$\lambda_m{=}0.8$ w/o shared backbone & 0.62 & 0.56 & 5.55 & 0.99 & 0.8669 & 0.2971 & 11.33 & 0.57 \\
$\lambda_m{=}0.8$ & 0.63 & \underline{0.58} & 5.56 & \underline{1.01} & 0.8674 & 0.2983 & \underline{11.47} & \underline{0.58} \\
\quad $+$ path/flow-product mixing ($\lambda{=}0.7$) & 0.65 & 0.57 & 5.56 & 0.96 & \underline{0.8687} & 0.2950 & 11.39 & \textbf{0.59} \\
\quad $+$ field regularization ($\lambda_r{=}1$) & \underline{0.66} & \underline{0.58} & \underline{5.58} & 0.96 & 0.8682 & 0.2952 & 11.44 & 0.57 \\
\quad $+$ regularization-corrected critic & \textbf{0.67} & \textbf{0.59} & \textbf{5.65} & \textbf{1.02} & \textbf{0.8702} & \textbf{0.3010} & \textbf{11.89} & \underline{0.58} \\
\bottomrule
\end{tabular}}
\end{table*}

\spara{Transition-Aware Field Estimation} This enables accurate tracking of the student field and notably improves instruction following (GenEval), text rendering (OCR), and diversity (LPIPS).

\section{Related Works}

\spara{Training-Free Diffusion Acceleration} Training-free methods for accelerating diffusion model sampling have progressed substantially~\citep{lu2022dpm, zhao2023unipc, xue2024accelerating, si2024freeu, ma2024surprising, li2025self}. However, diffusion distillation remains essential for high-quality few-step generation.

\spara{Diffusion Distillation} Distilled models generate images from noise in one or a few steps. Two main approaches transfer knowledge from pre-trained DMs:
1) Trajectory distillation~\citep{luhman2021knowledge,on_distill,salimans2021progressive,song2023consistency,song2024improved,yan2024perflow} matches teacher ODE trajectories at the instance level, facing challenging instance-wise alignment and numerical errors in teacher PF-ODE solutions;
2) Distribution matching via score distillation~\citep{yin2023one,luo2023diff,sid, sida,dpdmd,ddmd} aligns student and teacher distributions using objectives such as reverse KL, Fisher divergence, and moment matching.
FGM aligns velocity fields but considers only Gaussian forward processes~\citep{fgm}, where its objective reduces to Fisher divergence~\citep{sid}. TDM~\citep{luo2025learning} has shown strong few-step performance by learning partial denoising at each step.
Our VA-TDM unifies score distillation, Fisher divergence, and moment matching while yielding new distillation objectives. Combined with transition-aware online field learning, it achieves state-of-the-art few-step performance.

\spara{Generative adversarial networks.} Early GAN research achieved strong results on domain-specific datasets~\citep{reed2016generative,zhang2017stackgan,karras2020analyzing,karras2021alias}, while it has been effectively extended these successes to ImageNet~\citep{brock2018large, sauer2022stylegan, russakovsky2015imagenet}. However, training instability and limited scalability remain key challenges. Large-scale text-to-image GANs still rely on convolutional architectures with complex designs~\citep{kang2023scaling,zhu2025exploring}, and scaling transformer-based GANs remains difficult~\citep{jiang2021transgan,lee2021vitgan}. Adversarial Flow Models (AFM)~\citep{afm} train many-step flow-matching models through an adversarial game in velocity space, requiring discriminator Jacobian-vector products (JVPs). Our work also formulates an adversarial game in velocity space, but targets few-step distillation and avoids expensive JVP computation.

\section{Conclusions}
We introduced Velocity Adversarial Distillation (VAD), a unified framework that guides distillation through velocity fields constructed from the student's sampling process.
By separating the choice of field from the generator update, VAD connects score distillation, Fisher-based matching, and moment matching, clarifies MMD's heuristic updates, and derives new training objectives.
Building on this framework, VA-TDM combines dual-form supervision with transition-aware online velocity learning to supervise partial denoising along few-step student trajectories, without adding inference cost. Experiments across text-to-image and text-to-video generation demonstrate VA-TDM consistently achieves state-of-the-art four-step performance. VA-TDM also scales effectively to 33B MiniMax-H3, outperforming Fast-H3 in human evaluation and substantially narrowing the perceptual quality gap to the many-step teacher. These results highlight the value of our work, establishing VAD as a principled foundation for designing effective few-step distillation methods.

\endgroup

\bibliography{references}
\bibliographystyle{plainnat}

\clearpage
\appendix
\crefalias{section}{appendix}
\crefalias{subsection}{subappendix}
\crefalias{subsubsection}{subsubappendix}
\section{Proofs of Main Results}
\label{app:vad_optimality}
\label{app:vad_equilibrium_assumptions}

Throughout, time variables and conditioning information are fixed.
For the population game in \cref{eq:vad_alternating_game}, assume
that the teacher field and sampled target have finite second moments,
and that critic optimization ranges over all square-integrable fields
under the student's marginal law. \cref{thm:vad_matching_equilibrium}
additionally assumes realizability: there exists a student $\theta^\star$
with $\bar\bv^F_{\theta^\star,\tau}=\bu_\tau$ almost everywhere under
$p_{\theta^\star,\tau}$. The critic maximizes $V_F$, while the student
minimizes $\mathcal L_{\mathrm{path}}^F$ with the critic fixed.

\subsection{Proof of \texorpdfstring{\cref{lem:vad_optimal_witness}}{Lemma \getrefnumber{lem:vad_optimal_witness}}}
\label{app:vad_optimal_witness_proof}

\begin{proof}
Conditioning on $\bx_\tau$ gives the marginal payoff
\begin{equation}
\label{eq:vad_optimality_marginal_payoff}
V_F(\bff,\theta)
=\mathbb E_{p_{\theta,\tau}}\left[
\bff^\top\bigl(\bu_\tau-\bar\bv^F_{\theta,\tau}\bigr)
-\frac12\|\bff\|^2\right].
\end{equation}
Writing $\br_\theta=\bu_\tau-\bar\bv^F_{\theta,\tau}$ and completing the
square yields
\begin{equation}
\label{eq:vad_optimality_square}
V_F(\bff,\theta)
=\frac12\mathbb E_{p_{\theta,\tau}}\|\br_\theta\|^2
-\frac12\mathbb E_{p_{\theta,\tau}}\|\bff-\br_\theta\|^2.
\end{equation}
The upper bound is attained exactly when $\bff=\br_\theta$ almost
everywhere under $p_{\theta,\tau}$, proving the optimal critic and value.
\end{proof}
Parameterizing $\bff_\psi^F=\bu_\tau-\mathbf h_\psi^F$ makes critic
optimization equivalent to the field regression in
\cref{eq:vad_field_training}, by the conditional variance decomposition.

\subsection{Proof of \texorpdfstring{\cref{thm:vad_matching_equilibrium}}{Proposition \getrefnumber{thm:vad_matching_equilibrium}}}
\label{app:vad_matching_equilibrium_proof}

\begin{proof}
Choose a realizable matching student $\theta^\star$ and the globally
zero critic $\bff^\star\equiv\mathbf0$. For any admissible critic,
\begin{equation}
V_F(\bff,\theta^\star)
=-\frac12\mathbb E_{p_{\theta^\star,\tau}}\|\bff\|^2
\leq0=V_F(\bff^\star,\theta^\star),
\end{equation}
so $\bff^\star$ is a best response. With this critic,
$\mathcal L_{\mathrm{path}}^F(\theta;\bff^\star)=0$ for every student,
so $\theta^\star$ is also a best response. Thus
$(\theta^\star,\bff^\star)$ is a Nash equilibrium. By
\cref{lem:vad_optimal_witness}, the maximized critic payoff is
nonnegative and vanishes at $\theta^\star$, establishing the claimed optimality.
\end{proof}

\subsection{Proof of \texorpdfstring{\cref{lem:vad_flow_product}}{Lemma \getrefnumber{lem:vad_flow_product}}}
\label{app:vad_flow_product_proof}

Our proof uses the conditional-projection principle underlying denoising
score matching~\citep{vincent2011connection} and the score-projection
identity used in SiD~\citep{sid} and SIM~\citep{luo2024onestep}.
FGM~\citep{fgm} develops the corresponding flow-product identity.
Here the argument applies to conditional target fields under the
student's composite sampling law, without requiring a Gaussian forward
kernel.

Assume differentiable marginal densities and conditional fields near
$\theta_0$, integrability, and domination sufficient to interchange
differentiation and integration. Reparameterized derivatives are assumed
to exist and commute with expectation. Input laws and time variables are
held fixed during differentiation; auxiliary parameters are frozen while
their input derivatives are retained.

\begin{proof}
Let $P_\theta$ be the joint sampling law, write $X=\bx_\tau$ and
$V_\theta^F=\bv_\tau^F(\bx_\tau\mid\bx_s,\hat\bx_0)$, and define
$\bar\bv_{\theta,\tau}^F(\bx)=\mathbb E_{P_\theta}[V_\theta^F\mid X=\bx]$.
For any admissible test field $\mathbf a_\theta$, conditional expectation gives
\begin{equation}
\mathbb E_{P_\theta}\left[
\mathbf a_\theta(X)^\top
\bigl(V_\theta^F-\bar\bv_{\theta,\tau}^F(X)\bigr)\right]=0.
\end{equation}
Fix $\theta_0$, set $\mathbf h_0^F=\bar\bv_{\theta_0,\tau}^F$, and hold
$\bff_0$ and $\mathbf h_0^F$ fixed as functions. The projection identity gives
\begin{align}
H(\theta;\theta_0)
&=\mathbb E_{P_\theta}\left[
\bff_0(X)^\top\bigl(V_\theta^F-\mathbf h_0^F(X)\bigr)\right]\\
&=\int p_{\theta,\tau}(\bx)\,
\bff_0(\bx)^\top
\bigl(\bar\bv_{\theta,\tau}^F(\bx)-\mathbf h_0^F(\bx)\bigr)\,\mathrm d\bx.
\end{align}
At $\theta=\theta_0$, the centered difference vanishes pointwise, so
its product with the sampling-density derivative is zero. Hence
\begin{equation}
\label{app:vad_product}
\left.\nabla_\theta H(\theta;\theta_0)\right|_{\theta_0}
=\mathbb E_{p_{\theta_0,\tau}}\left[
\bff_0(\bx)^\top
\left.\partial_\theta\bar\bv_{\theta,\tau}^F(\bx)\right|_{\theta_0}\right],
\end{equation}
where $\partial_\theta$ differentiates at a fixed spatial argument.
Since $\mathcal G_{\mathrm{fp}}^F=-H$, this is precisely
\begin{equation}
\left.\nabla_\theta\mathcal G_{\mathrm{fp}}^F\right|_{\theta_0}
=\left.\nabla_\theta
\mathcal G_{\mathrm{mean}}^F(\theta;\theta_0,\bff_0)\right|_{\theta_0},
\end{equation}
whose right-hand side holds the outer measure fixed at $p_{\theta_0,\tau}$.
The argument allows both the target and the joint law to depend on
$\theta$, and applies to both A-form and B-form.
\end{proof}
The practical mixed update in \cref{eq:vad_mixed_loss} uses learned
auxiliary fields and detaches the critic output while retaining input
gradients through the centering field.

\subsection{Proof of \texorpdfstring{\cref{thm:vad_unified_connections}}{Theorem \getrefnumber{thm:vad_unified_connections}}}
\label{app:vad_unified_connections_proof}
\label{app:vad_mmd_connection}

Fix $0\leq s<t\leq1$, $s<\tau<1$, and $\eta=1$. Let $P_\theta$ denote
the law of the local reparameterized chain
\begin{equation}
\label{app:vad_chain}
\hat\bx_0=g_\theta(\bx_t,t),\qquad
\bx_s=\frac{s}{t}\bx_t+\left(1-\frac{s}{t}\right)\hat\bx_0,\qquad
\bx_\tau=\alpha_{\tau\mid s}\bx_s+\beta_{\tau\mid s}\beps,
\end{equation}
with fixed input law and independent standard Gaussian $\beps$.
Use the regularity conditions of \cref{app:vad_flow_product_proof},
an exact teacher, and finite reverse KL with the boundary conditions
needed for its pathwise gradient. Write $\mathrm D_\theta$ for total
sample derivatives with base noises fixed, and $\partial_\theta$ for
field derivatives at fixed spatial arguments. All gradient identities are
evaluated at a reference student $\bar\theta$, using the exact reference
field $\bar\bv^F_{\bar\theta,\tau}$ and critic
$\bu_\tau-\bar\bv^F_{\bar\theta,\tau}$.
For a teacher-regularized regressor, the unregularized reference field is
$(1+\lambda_r)\mathbf h_{\psi^\star}^F-\lambda_r\bu_\tau$.

\begin{proof}
\textbf{(i) Matching.}
For A-form, conditioning the target on $\bx_\tau$ gives
\begin{equation}
\bar\bv^A_{\theta,\tau}(\bx)=\frac{\bx-\mathbf m_\theta(\bx)}{\tau},
\qquad
\bff_A^\star(\bx)=\frac{\mathbf m_\theta(\bx)-\mathbf m_q(\bx)}{\tau},
\end{equation}
with the moments defined in \cref{eq:vad_b_matching}. Thus zero
residual is equivalent to conditional-moment matching.

For B-form, Gaussian differentiation gives
$\mathbb E[\beps\mid\bx_\tau=\bx]
=-\beta_{\tau\mid s}\nabla\log p_{\theta,\tau}(\bx)$.
Using the coefficients of \cref{eq:vad_kernel}, with
$\alpha=\alpha_{\tau\mid s}$ and $\beta=\beta_{\tau\mid s}$, yields
\begin{equation}
\label{eq:vad_velocity_score}
\begin{aligned}
\bar\bv^B_{\theta,\tau}(\bx)
&=\frac{\dot\alpha}{\alpha}\bx
-\left(\beta\dot\beta-\frac{\dot\alpha}{\alpha}\beta^2\right)
\nabla\log p_{\theta,\tau}(\bx)\\
&=-\frac{\bx}{1-\tau}
-\frac{\tau}{1-\tau}\nabla\log p_{\theta,\tau}(\bx),\\
\bff_B^\star
&=\frac{\tau}{1-\tau}
(\nabla\log p_{\theta,\tau}-\nabla\log q_\tau).
\end{aligned}
\end{equation}
The kernel maps $q_s$ to $q_\tau$, since
$\alpha_{\tau\mid s}(1-s)=1-\tau$ and
$\alpha_{\tau\mid s}^2s^2+\beta_{\tau\mid s}^2=\tau^2$.
Gaussian smoothing gives positive densities on $\mathbb R^d$, so equal
scores imply equal normalized densities. Since $\alpha_{\tau\mid s}>0$
and the Gaussian characteristic function never vanishes, deconvolution
makes $p_{\theta,\tau}=q_\tau$ equivalent to $p_{\theta,s}=q_s$.

\textbf{(ii) The clean endpoint.}
At $s=0$, $\bx_s=\hat\bx_0$ and
$\bx_\tau=(1-\tau)\hat\bx_0+\tau\mathbf z_\tau$, giving
\begin{equation}
\bv_\tau^A=\frac{\bx_\tau-\hat\bx_0}{\tau}
=\mathbf z_\tau-\hat\bx_0=\bv_\tau^B.
\end{equation}
The sampled targets coincide, so their conditional means, optimal critics,
and gradients under identical differentiation conventions coincide.

\textbf{(iii) Pathwise updates with the critic output detached.}
Set $a=1-s/t>0$ and $c=\alpha_{\tau\mid s}a\in(0,1)$. From
\cref{app:vad_chain},
\begin{equation}
\label{app:vad_path_coefficients}
\mathrm D_\theta\bx_s=a\,\mathrm D_\theta\hat\bx_0,
\qquad
\mathrm D_\theta\bx_\tau=c\,\mathrm D_\theta\hat\bx_0,
\end{equation}
and the sampled targets satisfy
\begin{equation}
\mathrm D_\theta\bv_\tau^A
=-\frac{1-c}{\tau}\mathrm D_\theta\hat\bx_0,
\qquad
\mathrm D_\theta\bv_\tau^B
=-\frac{a}{1-s}\mathrm D_\theta\hat\bx_0.
\end{equation}
The reverse-KL pathwise identity is
\begin{equation}
\nabla_\theta\mathrm{KL}(p_{\theta,\tau}\|q_\tau)
=\mathbb E\left[(\mathrm D_\theta\bx_\tau)^\top
(\nabla\log p_{\theta,\tau}-\nabla\log q_\tau)(\bx_\tau)\right].
\end{equation}
Substituting the optimal critics into \cref{eq:vad_stopped_updates} gives
\begin{align}
\label{app:vad_pathwise_mmd}
\nabla_\theta\mathcal L_{\mathrm{path,sg}}^A
&=\frac{1-c}{\tau^2}\mathbb E\left[
(\mathrm D_\theta\hat\bx_0)^\top
(\mathbf m_\theta-\mathbf m_q)(\bx_\tau)\right],\\
\label{app:vad_pathwise_kl}
\nabla_\theta\mathcal L_{\mathrm{path,sg}}^B
&=\frac{a}{1-s}\mathbb E[(\mathrm D_\theta\hat\bx_0)^\top\bff_B^\star]
=\frac{\tau}{(1-\tau)^2}
\nabla_\theta\mathrm{KL}(p_{\theta,\tau}\|q_\tau).
\end{align}
The A-form expectation is the MMD surrogate update made explicit below;
the B-form gives reverse KL. Both coefficients are positive time weights.

\textbf{(iv) Flow-product updates without output detachment.}
Define the discrepancy with its outer measure held fixed:
\begin{equation}
\label{eq:vad_stopped_measure}
\mathcal D_F(\theta;\bar\theta)
=\frac12\mathbb E_{p_{\bar\theta,\tau}}
\|\bu_\tau-\bar\bv^F_{\theta,\tau}\|^2.
\end{equation}
\cref{lem:vad_flow_product}, applied to the optimal reference critic, gives
\begin{equation}
\left.\nabla_\theta\mathcal G_{\mathrm{fp}}^F\right|_{\bar\theta}
=-\mathbb E_{p_{\bar\theta,\tau}}\left[
\left(\left.\partial_\theta\bar\bv^F_{\theta,\tau}\right|_{\bar\theta}\right)^\top
\bff_F^\star\right]
=\left.\nabla_\theta\mathcal D_F(\theta;\bar\theta)\right|_{\bar\theta}.
\end{equation}
The residual identities in part (i) identify the objectives as
\begin{align}
\mathcal D_A(\theta;\bar\theta)
&=\frac1{2\tau^2}\mathbb E_{p_{\bar\theta,\tau}}
\|\mathbf m_\theta-\mathbf m_q\|^2,\\
\mathcal D_B(\theta;\bar\theta)
&=\frac{\tau^2}{2(1-\tau)^2}\mathbb E_{p_{\bar\theta,\tau}}
\|\nabla\log p_{\theta,\tau}-\nabla\log q_\tau\|^2.
\end{align}
These are the conditional-moment and scaled Fisher discrepancies,
respectively, with the fixed-outer-measure gradient convention.

\textbf{MMD objective and surrogate.}
To make the MMD correspondence explicit~\citep{mmd},
write $\boldsymbol\delta_\theta=\mathbf m_\theta-\mathbf m_q$.
Its moment objective and local fixed-measure version are
\begin{equation}
\label{eq:vad_mmd_objective}
J_{\mathrm{MM}}(\theta)=\frac12\mathbb E_{p_{\theta,\tau}}
\|\boldsymbol\delta_\theta\|^2,
\qquad
L_{\mathrm{MM}}(\theta;\bar\theta)
=\frac12\mathbb E_{p_{\bar\theta,\tau}}
\|\boldsymbol\delta_\theta\|^2=\tau^2\mathcal D_A(\theta;\bar\theta).
\end{equation}
The exact local gradient is
\begin{equation}
\label{eq:vad_mmd_exact_gradient}
\mathbf g_{\mathrm{MM}}
=\left.\nabla_\theta L_{\mathrm{MM}}\right|_{\bar\theta}
=\mathbb E_{p_{\bar\theta,\tau}}\left[
\left(\left.\partial_\theta\mathbf m_\theta\right|_{\bar\theta}\right)^\top
\boldsymbol\delta_{\bar\theta}\right].
\end{equation}
MMD approximates the implicit conditional-mean derivative by the sampled
clean-prediction derivative, giving
\begin{equation}
\label{eq:vad_mmd_surrogate_gradient}
\mathbf g_{\mathrm{MM}}\approx\mathbf g_{\mathrm{sur}}
\coloneqq\mathbb E_{P_{\bar\theta}}\left[
\left(\left.\mathrm D_\theta\hat\bx_0\right|_{\bar\theta}\right)^\top
\boldsymbol\delta_{\bar\theta}(\bx_\tau)\right].
\end{equation}
This is the gradient of the detached surrogate
\begin{equation}
\label{eq:vad_mmd_surrogate}
L_{\mathrm{sur}}(\theta)
=\mathbb E_{P_\theta}\left[
\hat\bx_0^\top\operatorname{sg}\!\left(
\boldsymbol\delta_\theta(\bx_\tau)\right)\right].
\end{equation}
Parts (iii)--(iv) therefore give both correspondences directly:
\begin{align}
\label{eq:vad_mmd_flow_product_gradient}
\tau^2\left.\nabla_\theta\mathcal G_{\mathrm{fp}}^A\right|_{\bar\theta}
&=\mathbf g_{\mathrm{MM}},\\
\label{eq:vad_mmd_pathwise_gradient}
\left.\nabla_\theta\mathcal L_{\mathrm{path,sg}}^A\right|_{\bar\theta}
&=\frac{1-c}{\tau^2}\mathbf g_{\mathrm{sur}}.
\end{align}
Thus flow-product recovers the exact local moment gradient, while the
detached pathwise update recovers MMD's surrogate direction up to time weighting.
\end{proof}

\section{Supporting Derivations}
\label{app:vad_velocity_construction}

\subsection{Reference DSM Variance}
\label{app:vad_variance_motivation}

Consider the reference Gaussian forward process with $0<s<\tau<1$,
$\bx_s=(1-s)\bx_0+s\mathbf e_s$, and
$\bx_\tau=\alpha_{\tau\mid s}\bx_s+\beta_{\tau\mid s}\mathbf e_\tau$,
where the standard Gaussian noises are independent of each other and
$\bx_0$. The two DSM targets
\begin{equation}
\mathbf S_{s\to\tau}
=-\frac{\bx_\tau-\alpha_{\tau\mid s}\bx_s}{\beta_{\tau\mid s}^2},
\qquad
\mathbf S_{0\to\tau}
=-\frac{\bx_\tau-(1-\tau)\bx_0}{\tau^2}
\end{equation}
have the same conditional mean $\nabla\log q_\tau(\bx_\tau)$.
The Gaussian bridge and the law of total covariance give the comparison
of \citet{rl_awm} in our notation:
\begin{equation}
\label{eq:vad_noisy_score_variance}
\operatorname{Cov}(\mathbf S_{s\to\tau}\mid\bx_\tau)
=\operatorname{Cov}(\mathbf S_{0\to\tau}\mid\bx_\tau)
+\left(\frac1{\beta_{\tau\mid s}^2}-\frac1{\tau^2}\right)\mathbf I.
\end{equation}
The excess is positive for $0<s<\tau<1$ and vanishes at $s=0$.
This reference-process comparison motivates the noisy-target DSM discussion
in \cref{sec:vad_targets}.

\subsection{Clean-Target and Path-Marginal Score Mismatch}
\label{app:vad_score_mismatch}

Fix $t,s,\tau$ with $s<\tau<1$ and the input law. Let
$R_\theta(\mathrm d\bx_s,\mathrm d\hat\bx_0)$ be the joint law of the
intermediate output and clean anchor. The two perturbed marginals are
\begin{equation}
\label{eq:vad_clean_perturbed_marginal}
\begin{aligned}
q(\bx_\tau\mid\hat\bx_0)
&=\mathcal N(\bx_\tau;(1-\tau)\hat\bx_0,\tau^2\mathbf I),\\
p_{\theta,\tau}^{\mathrm{clean}}(\bx_\tau)
&=\int q(\bx_\tau\mid\hat\bx_0)
R_\theta(\mathrm d\bx_s,\mathrm d\hat\bx_0),\\
p_{\theta,\tau}^{\mathrm{path}}(\bx_\tau)
&=\int q(\bx_\tau\mid\bx_s)
R_\theta(\mathrm d\bx_s,\mathrm d\hat\bx_0).
\end{aligned}
\end{equation}
TDM's ideal importance-weighted clean-target objective~\citep{luo2025learning}
uses the ratio of these conditional kernels. Since both kernels have
positive densities, for any integrable loss $\ell_\psi$,
\begin{align}
\label{eq:vad_tdm_importance_weight}
w(\bx_\tau,\bx_s,\hat\bx_0)
&=\frac{q(\bx_\tau\mid\hat\bx_0)}{q(\bx_\tau\mid\bx_s)},\\
\label{eq:vad_importance_identity}
\mathbb E_{R_\theta q(\bx_\tau\mid\bx_s)}[w\ell_\psi]
&=\mathbb E_{R_\theta q(\bx_\tau\mid\hat\bx_0)}[\ell_\psi].
\end{align}
For $\ell_\psi=\omega_\tau\|h_\psi(\bx_\tau,\tau)-\hat\bx_0\|^2$
with $\omega_\tau>0$, an unrestricted regressor with finite second moments
has optimum
$h^*(\bx_\tau,\tau)=\mathbb E_{R_\theta q(\bx_\tau\mid\hat\bx_0)}
[\hat\bx_0\mid\bx_\tau]$.
Gaussian differentiation identifies its score and the required path score as
\begin{equation}
\label{eq:vad_clean_score_optimum}
\begin{aligned}
\nabla\log p_{\theta,\tau}^{\mathrm{clean}}(\bx_\tau)
&=-\frac{\bx_\tau-(1-\tau)h^*(\bx_\tau,\tau)}{\tau^2},\\
\nabla\log p_{\theta,\tau}^{\mathrm{path}}(\bx_\tau)
&=\mathbb E_{R_\theta q(\bx_\tau\mid\bx_s)}\left[
-\frac{\bx_\tau-\alpha_{\tau\mid s}\bx_s}{\beta_{\tau\mid s}^2}
\,\middle|\,\bx_\tau\right].
\end{aligned}
\end{equation}
Thus importance weighting recovers the clean-denoising objective, whose
population score can differ from the path score needed by the generator.
For example, in one dimension with $\hat x_0=\hat\epsilon=0$ and
$\eta=1$, the transition gives $x_s=0$: the path marginal is
$\mathcal N(0,\beta_{\tau\mid s}^2)$ and the clean marginal is
$\mathcal N(0,\tau^2)$, with different variances for $s>0$.
At $s=0$, $\bx_s=\hat\bx_0$ and the two marginals coincide.

\newcommand{\expdetailblank}[1][0.8cm]{\mbox{\rule{#1}{0.35pt}}}

\section{Experimental Details}
\label{app:experimental_details}

This section provides the training details for the experiments
in \cref{sec:vatdm_experiments}.

\subsection{Training Configurations}
\label{app:training_details}

VA-TDM trains a generator and an online auxiliary field while keeping the
teacher fixed. The generator combines the two velocity forms through
\cref{eq:vad_dual_form_loss}, and the auxiliary field follows the
regression and regularization objectives in
\cref{sec:vad_trajectory}. By default, we apply LoRA fine-tuning with rank 64. We update the online score and generator once each iteration. We use the AdamW optimizer and set $(\beta_1,\beta_2)=(0.,0.95)$.

\subsection{User Study}
\label{app:h3_user_study_details}

The study in \cref{fig:h3_user_study} compares MiniMax-H3 with Fast-H3,
VA-TDM with MiniMax-H3, and VA-TDM with Fast-H3. We carried out a user study in which participants were shown two anonymized audio-videos, each produced by a different model, and were asked to pick the one with superior audio-visual quality and closer alignment with the prompt. For audio-video generation, we randomly chose 10 prompts, and every generated audio-video was manually checked to confirm it contained no inappropriate or harmful content.

\begingroup
\makeatletter
\setlength{\@fptop}{0pt}
\setlength{\@fpsep}{18pt}
\makeatother
\section{Additional Visual Samples}
\label{app:additional_visual_samples}

\crefrange{fig:h3_additional_1}{fig:h3_additional_5} provide
additional text-to-video comparisons complementing \cref{fig:teaser}.
Each figure presents six selected frames per method, with the summarized
prompt shown above the panels. From top to bottom, the rows correspond to
MiniMax-H3 (100 NFE), Fast-H3 (4 NFE), and VA-TDM-H3 (4 NFE).
The examples cover human--animal interactions, clothing changes, fantasy
combat, stylized game scenes, and multi-scene cinematic sequences.

\begin{figure}[!htbp]
    \centering
    \includegraphics[width=1.0\linewidth]{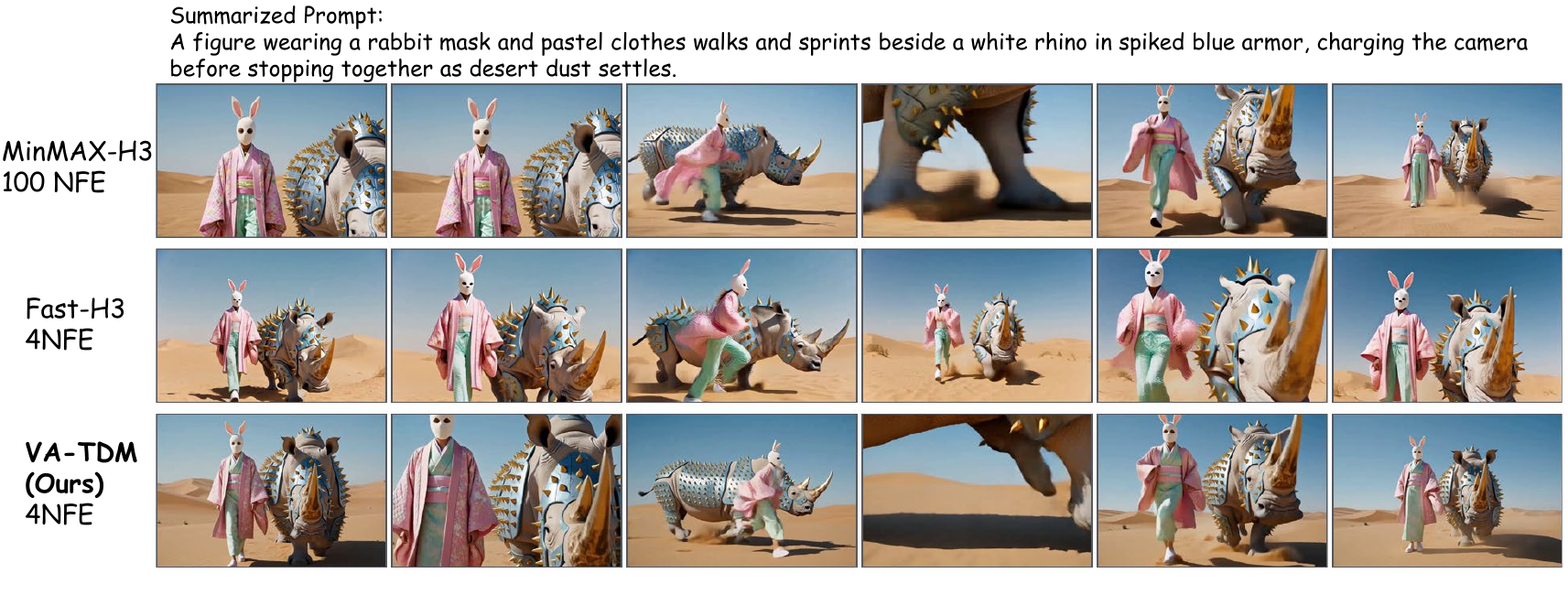}
    \caption{Additional visual comparison of MiniMax-H3 (100 NFE), Fast-H3
    (4 NFE), and VA-TDM-H3 (4 NFE).}
    \label{fig:h3_additional_1}
\end{figure}

\begin{figure}[!htbp]
    \centering
    \includegraphics[width=1.0\linewidth]{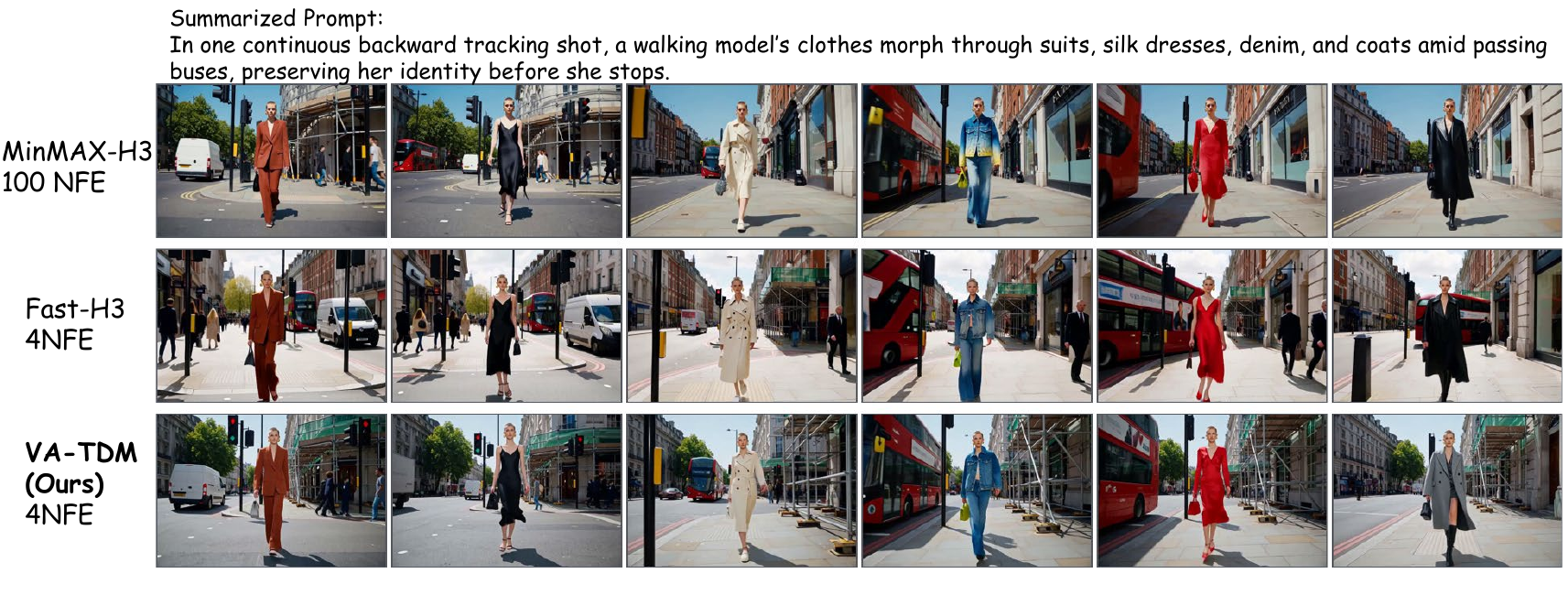}
    \caption{Additional visual comparison of MiniMax-H3 (100 NFE), Fast-H3
    (4 NFE), and VA-TDM-H3 (4 NFE).}
    \label{fig:h3_additional_2}
\end{figure}

\begin{figure}[!htbp]
    \centering
    \includegraphics[width=1.0\linewidth]{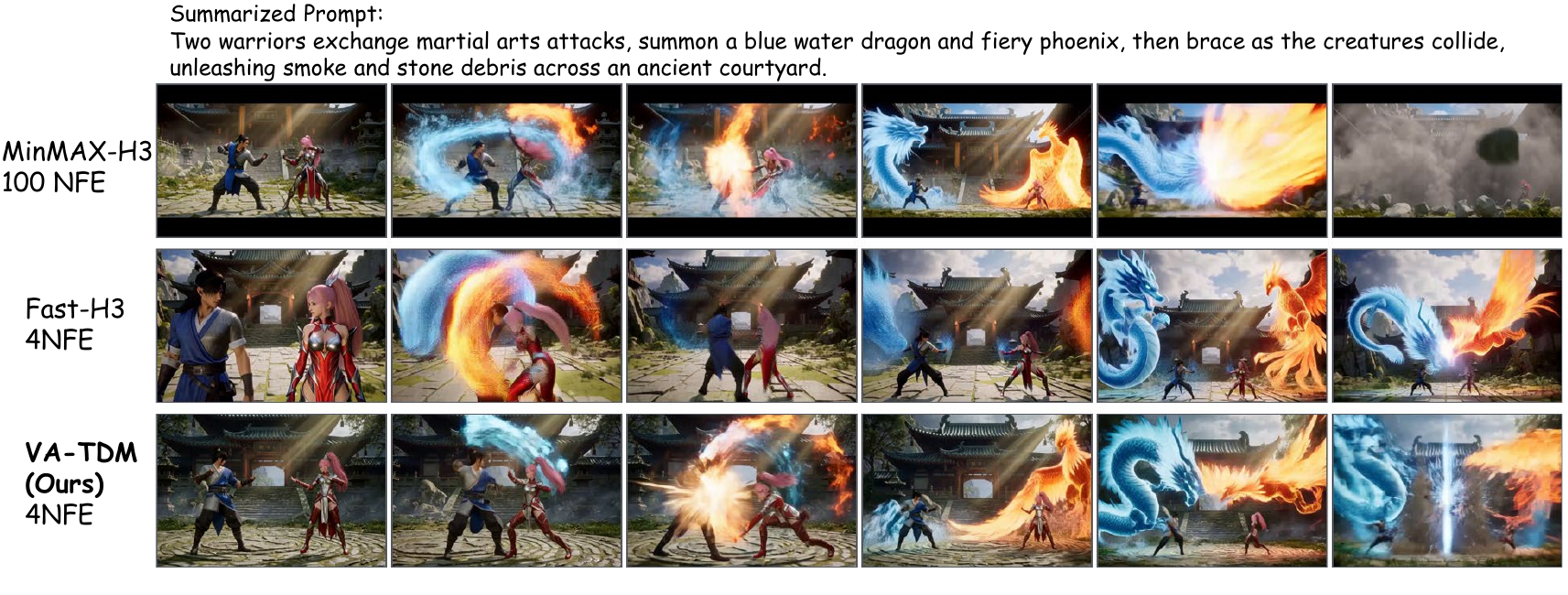}
    \caption{Additional visual comparison of MiniMax-H3 (100 NFE), Fast-H3
    (4 NFE), and VA-TDM-H3 (4 NFE).}
    \label{fig:h3_additional_3}
\end{figure}

\clearpage
\begin{figure}[!htbp]
    \centering
    \includegraphics[width=1.0\linewidth]{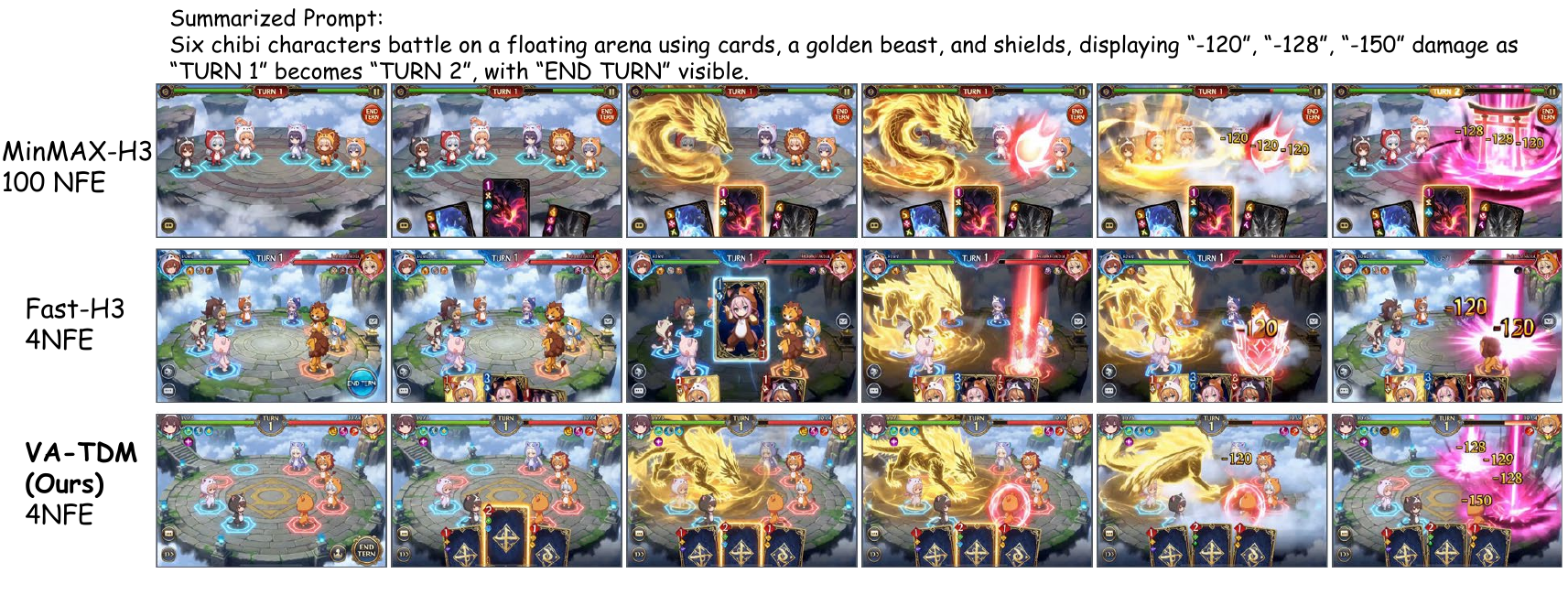}
    \caption{Additional visual comparison of MiniMax-H3 (100 NFE), Fast-H3
    (4 NFE), and VA-TDM-H3 (4 NFE).}
    \label{fig:h3_additional_4}
\end{figure}

\begin{figure}[!htbp]
    \centering
    \includegraphics[width=1.0\linewidth]{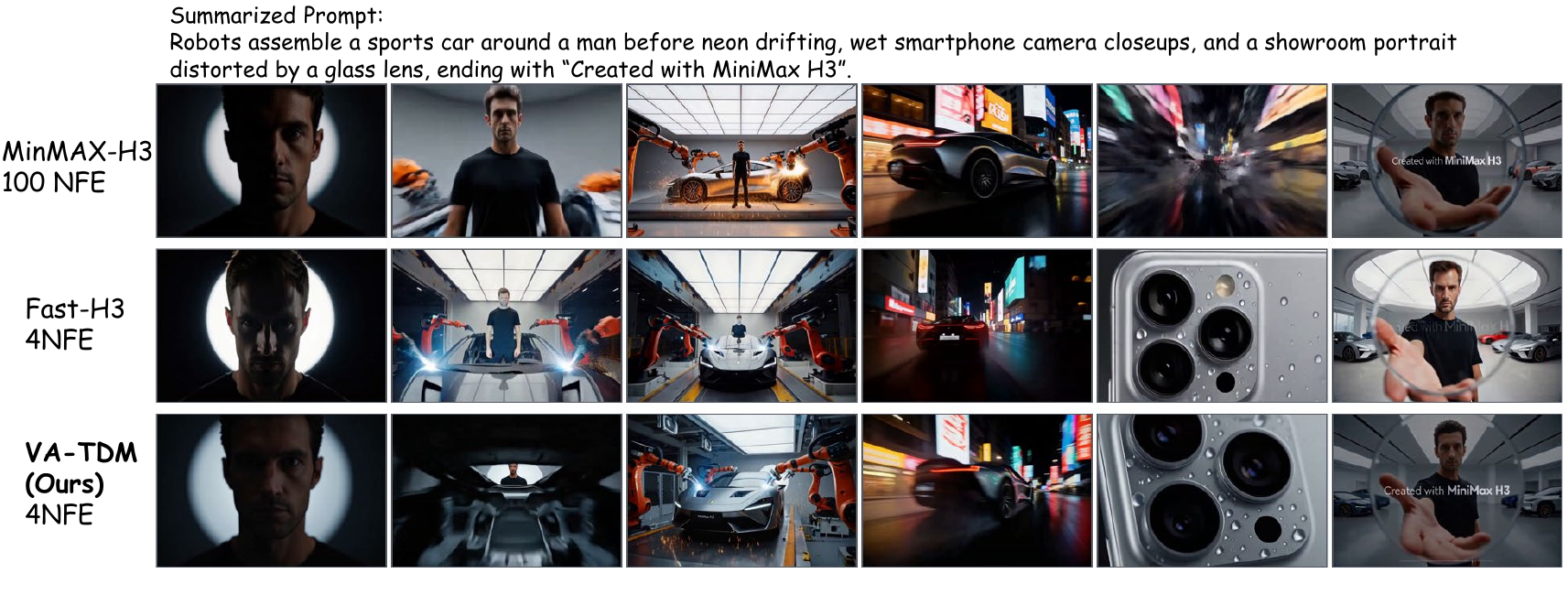}
    \caption{Additional visual comparison of MiniMax-H3 (100 NFE), Fast-H3
    (4 NFE), and VA-TDM-H3 (4 NFE).}
    \label{fig:h3_additional_5}
\end{figure}
\clearpage
\endgroup

\end{document}